\documentclass[10pt,journal]{IEEEtran}
\usepackage{eqnarray,amsmath}
\usepackage{amssymb}
\usepackage{array}
\usepackage{bm}
\usepackage{multirow}
\usepackage{graphicx} 
\usepackage{overpic}
\usepackage{epstopdf}
\usepackage{subfigure}
\usepackage{url}
\usepackage{algorithmicx}
\usepackage[ruled]{algorithm}
\usepackage{amsthm} 
\usepackage{bm}
\usepackage{enumerate}
\usepackage{algpseudocode}
\usepackage{times}
\usepackage{color}
\usepackage{multirow}

\newtheorem{Lemma}{Lemma}
\newtheorem{Proposition}{Proposition}
\newtheorem{Theorem}{Theorem}
\newtheorem{Definition}{Definition}

\newcommand{\tabincell}[2]{\begin{tabular}{@{}#1@{}}#2\end{tabular}}

\ifCLASSINFOpdf
\else
\fi
\begin{document}
%
\title{Learning Support Correlation Filters\\ for Visual Tracking}
%
%
%

\author{Wangmeng Zuo, 
        Xiaohe Wu, 
        Liang Lin, 
        Lei Zhang,  
        and Ming-Hsuan Yang 
}



\IEEEcompsoctitleabstractindextext{%
\begin{abstract}
Sampling and budgeting training examples
are two essential factors in tracking algorithms based on support vector machines (SVMs)
as a tradeoff between accuracy and efficiency.
Recently, the circulant matrix formed by
dense sampling of translated image patches
has been utilized in correlation filters for fast tracking.
%
%
In this paper, we derive an equivalent formulation of a SVM model with circulant matrix
expression and present an efficient alternating optimization method for visual tracking.
We incorporate the discrete Fourier transform with the proposed
alternating optimization process, and pose the tracking problem
as an iterative learning of support correlation filters (SCFs)
which find the global optimal solution with real-time performance.
%
%
For a given circulant data matrix with $n^2$ samples of size $n \times n$,
the computational complexity of the proposed algorithm is $O(n^2 \log n)$
whereas that of the standard SVM-based approaches is at least $O(n^4)$.
In addition, we extend the SCF-based tracking algorithm with multi-channel features,
%
kernel functions, and scale-adaptive approaches to further improve the tracking performance.
Experimental results on a large benchmark dataset show that the proposed SCF-based algorithms
perform favorably against the state-of-the-art tracking methods in terms of accuracy and speed.
\end{abstract}
}

\maketitle

\IEEEdisplaynotcompsoctitleabstractindextext
\IEEEpeerreviewmaketitle


\section{Introduction}
\IEEEPARstart{R}{obust} visual tracking is a challenging problem
due to the large changes of object appearance caused by pose, illumination, deformation, occlusion,
distractors, as well as background clutters.
Among the state-of-the-art methods, discriminative classifiers with model update and
dense sampling have been demonstrated to perform well in visual tracking.
%
On the other hand, correlation filters have been shown to be efficient for locating objects
with the use of circulant matrix and fast Fourier transform.
%
%
%
Central to the advances in visual tracking
are the development of effective appearance models and efficient sampling schemes.

Discriminative appearance models have been extensively studied in visual tracking
and have achieved the state-of-the-art results.
One representative discriminative appearance model is based on support vector machines (SVMs) \cite{avidan2004support,bai2012robust,hare2011struck,zhang2014meem}.
To learn classifiers for detecting objects within local regions,
SVM-based tracking approaches are developed based on
two modules: a {\em sampler} to generate a set of positive and negative samples and a {\em learner} to update the classifier using the training samples.
To reduce the computational load, SVM-based trackers usually only use
a limited set of samples \cite{hare2011struck,zhang2014meem}.
As kernel SVM-based tracking methods are susceptible to the {\em curse of kernelization},
a budget mechanism is introduced for online learning of the structural SVM tracker \cite{hare2011struck}
to restrict the number of support vectors,
or an explicit feature mapping function
is used to approximate the intersection kernel \cite{zhang2014meem}.
While sampling and budgeting may improve tracking efficiency at the expense of accuracy,
most SVM-based trackers \cite{bai2012robust,hare2011struck,zhang2014meem}
do not run in real-time.

Correlation filters (CFs) \cite{bolme2010visual, henriques2015high, henriques2012exploiting, zhang2014fast} have recently been utilized for efficient visual tracking.
%
The data matrix formed by dense sampling of base sample has circulant structures,
which facilitates the use of the discrete Fourier transform (DFT) for efficient and
effective visual tracking \cite{bolme2010visual, henriques2015high, henriques2012exploiting, zhang2014fast}.
%
However, ridge regression or kernel ridge regression are generally adopted as the predictors in these trackers.
Henriques \textit{et al.} \cite{henriques2013beyond} apply the circulant property for
training of support vector regression efficiently to detect pedestrians.
%
The problem on how to exploit the circulant property to accelerate SVM-based trackers
remains unaddressed.

In this paper, we propose a novel SVM-based algorithm via
support correlation filters (SCFs) for efficient and accurate visual tracking.
Different from the existing SVM-based trackers, the proposed algorithm based on SCFs
deals with the sampling and budgeting issues by using the data matrix formed by
%
dense sampling.
By exploiting the circulant property, we formulate the proposed
SVM-based tracker as a learning problem for support correlation filters
and propose an efficient algorithm.
By incorporating the discrete Fourier transform in an alternating optimization process,
the SVM classifier can be efficiently updated by iterative learning of correlation filters.
For an $n \times n$ image, there are $n^2$ training sample images of the same size
in the circulant data matrix and the computational complexity of the proposed algorithm is
$O(n^2 \log n)$ whereas that of the standard SVM-based approaches is at least $O(n^4)$.
Furthermore, we extend the proposed SCF-based algorithm
to multi-channel SCF (MSCF), kernelized SCF (KSCF),
%
and scale-adaptive KSCF (SKSCF) methods to improve the tracking performance.

We evaluate the proposed SCF-based algorithms
on a large benchmark dataset with comparison to
the state-of-the-art methods \cite{wu2013online} and analyze the tracking results.
First, with the discriminative strength of SVMs, the proposed KSCF method
performs favorably against the existing regression-based correlation filter trackers.
Second, by exploiting the circulant structure of training samples, the proposed KSCF
algorithm performs well compared with the existing SVM-based trackers \cite{hare2011struck,zhang2014meem} in terms of efficiency and accuracy.
Third, the proposed KSCF and SKSCF algorithms
outperform the state-of-the-art methods including the ensemble
and scale-adaptive tracking methods \cite{zhang2014meem, danelljan2014accurate, li2014scale}.
%
\section{Related Work and Problem Context}
\label{relatedwork}
Visual tracking has long been an active research topic in computer vision which involves
developments of
both learning methods (e.g., feature learning and selection, online learning
and ensemble models)
and application domains (e.g., auto-navigation, visual surveillance and
human-computer interactions).
Several surveys and performance evaluation on state-of-the-art tracking algorithms \cite{pang2013finding,smeulders2014visual,wu2013online,yang2011recent} have been
reported in the literature,
and in this section we discuss the most relevant methods to this work.

\begin{figure*}[t]
\begin{center}
\includegraphics[width=\textwidth]{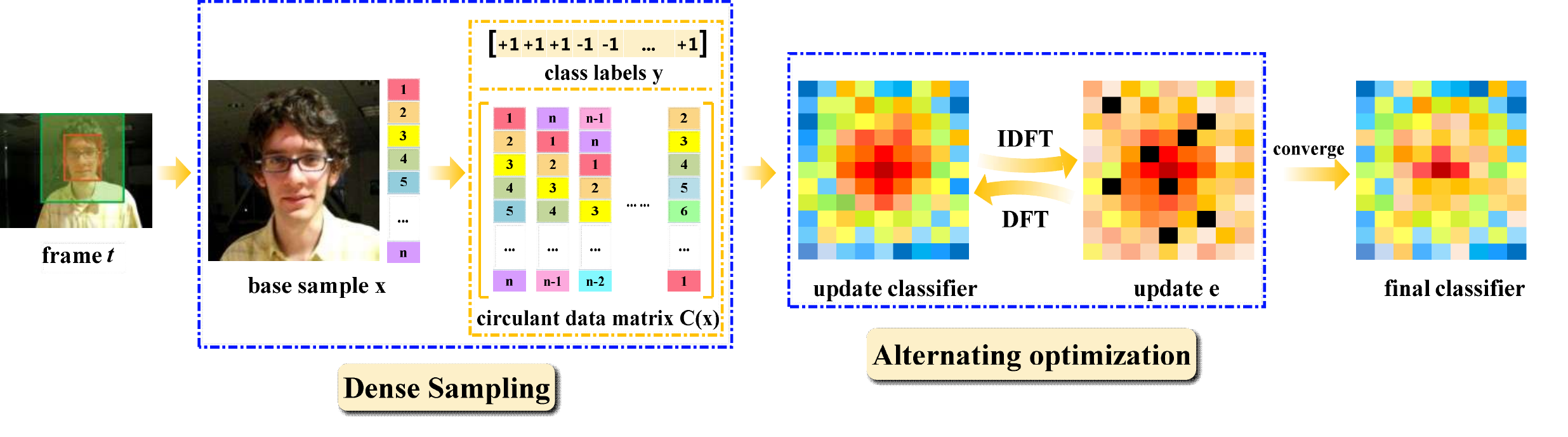}
\end{center}
\vspace{-0.1in}
 \caption{Illustration of the proposed SCF learning algorithm at the $t$-th frame.
 The proposed algorithm iterates between updating ${\bf e}$ and updating SVM classifier $\{ {\bf w}, b\}$ until convergence.
 In each iteration, only one DFT and one IDFT are required, which make
 the proposed algorithm computationally efficient.
The black blocks in ${\bf e}$ denote support vectors, and our algorithm can adaptively find and exploit difficult samples (i.e., support vectors) to learn support correlation filters.}
\label{fig:scf_learn}
\end{figure*}

\subsection{Appearance models for visual tracking}
Appearance models play an important role in visual tracking which can be
broadly categorized as generative or discriminative.
%
Generative appearance methods based on holistic templates \cite{zhong2012robust}, subspace representations \cite{hu2012single,lim2004incremental,ho2004visual},
and sparse representations \cite{jia2012visual,chen2011object,wang2015inverse} have been developed for object representations.
Discriminative appearance methods
are usually based on features learned from a large set of examples with effective classifiers.
Visual tracking is posed as a task to distinguish the target objects from the backgrounds.
%
%
Tracking methods based on discriminative appearance models
have been shown to achieve the state-of-the-art results \cite{wu2013online}.

Discriminative tracking methods are usually based on
object detection within local search using classifiers such as
boosting methods \cite{babenko2009visual, babenko2011robust, grabner2006real, grabner2008semi,kalal2012tracking},
random forests \cite{saffari2009line,santner2010prost}, and SVMs \cite{avidan2004support,bai2012robust, hare2011struck}.
Among these classifiers, boosting methods \cite{babenko2009visual, babenko2011robust,
grabner2006real, grabner2008semi, kalal2012tracking} and random forests
\cite{saffari2009line, santner2010prost} are ensemble learning methods
%
where sampling from large sets of features is indispensable,
and that makes it difficult to adapt correlation filters in these approaches.
In this work, we exploit the discriminative strength of SVMs and efficiency of
correlation filters for visual tracking.

Label ambiguity has also been studied for visual
tracking, e.g., semi-supervised  \cite{grabner2008semi, kalal2012tracking, tang2007co} and
multiple instance \cite{babenko2009visual, babenko2011robust}
learning methods.
Considering that classification based methods are trained to
predict the class label rather than the object location,
Hare \textit{et al.} \cite{hare2011struck} propose a tracker based on structured SVM.
In this work, we alleviate the label ambiguity problem by using the assignment scheme
in a way similar to that for object detection and tracking \cite{girshick2014rich, grabner2006real, uijlings2013selective}.

\subsection{Correlation filters for tracking}
A correlation filter uses a designed template to generate strong response to a region that is
similar to the target object while suppressing responses to distractors.
Correlation filters have been widely applied to numerous problems such as face recognition \cite{kumar2006correlation,savvides2003efficient}, object detection \cite{bolme2009average,henriques2013beyond,mahalanobis1987minimum}, object alignment \cite{boddeti2013correlation} and action recognition \cite{gondal2011action,Rodriguez2008action}.
%
%
A number of correlation filters have been proposed in the literature including the
%
minimum average correlation energy (MACF) \cite{mahalanobis1987minimum},
optimal trade-off synthetic discriminant filter (OTSDF) \cite{refregier1991optimal},
unconstrained minimum average correlation energy (UMACE) \cite{savvides2003efficient}, and
minimum output sum of squared error (MOSSE) \cite{bolme2010visual} methods.

%
Recently, the max-margin CF (MMCF) \cite{rodriguez2013maximum},
multi-channel CF \cite{danelljan2014accurate, danelljan2014adaptive, galoogahi2013multi, henriques2015high}, and
kernelized CF \cite{henriques2015high, henriques2012exploiting, patnaik2009fast} methods
have been developed for object detection and tracking.
The MMCF \cite{rodriguez2013maximum} scheme combines
the localization properties of correlation filters with good generalization performance of SVM.
The multi-channel correlation filters \cite{danelljan2014accurate, danelljan2014adaptive, galoogahi2013multi, henriques2015high} are designed to use more
effective features, e.g., histogram of oriented gradients (HOG).
In addition, a method that combines MMCF and multi-channel CF is developed \cite{boddeti2014maximum} for object detection and landmark localization.
The kernel tricks are also employed to learn kernelized synthetic discriminant functions (SDF)\cite{patnaik2009fast} with correlation filters.
We note that the MMCF \cite{boddeti2014maximum,rodriguez2013maximum} and kernelized SDF \cite{patnaik2009fast} schemes are trained off-line
with high computational load, and do not exploit the circulant structure of data matrix formed by
translated images of target objects.

In visual tracking, Bolme \textit{et al.} \cite{bolme2010visual} propose the MOSSE method
to learn adaptive correlation filters with high efficiency and competitive performance.
Subsequently, the kernelized correlation filter (KCF) \cite{henriques2015high} is
developed by exploiting the circulant property of the kernel matrix.
Extensions of CF and KCF with multi-channel features are introduced for visual tracking \cite{danelljan2014accurate,danelljan2014adaptive,galoogahi2013multi,henriques2015high}.
Within the tracking methods based on correlation filters,
numerous issues such as adaptive scale estimation \cite{danelljan2014accurate, li2014scale,zhang2014fast}, limited boundaries \cite{Galoogahi2015Correlation},
zero-aliasing \cite{fernandez2013zero}, tracking failure \cite{Ma2015Long}, and
partial occlusion \cite{Liu2015Real} have been addressed.

%
We note existing CF-based tracking methods are developed with ridge regression schemes for locating the target.
On the other hand, the
SVM-based tracking methods, e.g., Struck \cite{hare2011struck} and MEEM \cite{zhang2014meem}, have been demonstrated to achieve the state-of-the-art performance.
One straightforward extension is to integrate SVM-based trackers with the MMCF method\cite{rodriguez2013maximum}.
Nevertheless, the MMCF scheme is computationally prohibitive for real-time applications.
%
In this work, we develop novel discriminative tracking algorithms based on SVMs and
correlation filters that perform efficiently and effectively.

\section{Support Correlation Filtering}
\label{SCF}
We first present the problem formulation and propose an alternating optimization algorithm
to learn support correlation filters efficiently.
We then develop the MSCF, KSCF and SKSCF methods to learn
%
multi-channel, nonlinear and scale-adaptive correlation filters respectively
for robust visual tracking.

\subsection{Problem formulation}
Given an image $\bf{x}$, the full set of its translated versions
forms a circulant matrix $\bf{X}$ with several interesting properties \cite{gray2006toeplitz}, where
each row represents one possible observation of a target object (See Fig. \ref{fig:scf_learn}).
A circulant matrix consists of all possible cyclic translations of a target image, and tracking
is formulated as determining the most likely row.
In general, the eigenvectors of a
circulant matrix $\bf{X}$ are the base vectors $F$ of the discrete Fourier transform:
\begin{eqnarray} \label{eq:dft}
{\bf X} = F{\bf}^H \mbox{Diag} ({\bf \hat{x}}) F,
\end{eqnarray}
where
 $F{\bf}^H $ is the Hermitian transpose of $F$
and ${\bf \hat{x}} = \mathcal{F} ({\bf x})$ denotes the Fourier transform of ${\bf x}$.
In the following, we use $\mbox{Diag}(\cdot)$ to form a diagonal matrix from a vector,
and use $\mbox{diag}(\cdot)$ to return the diagonal vector of a matrix.

Our goal is to learn a support correlation filter ${\bf w}$
and a bias $b$, to classify any translated image ${\bf x}_i$ by
\begin{eqnarray} \label{eq:class_x}
y_i = \mbox{sgn} \left( {\bf w}^\top {\bf x}_i + b \right).
\end{eqnarray}
\noindent
Note that all the translated images ${\bf x}_i$ form a circulant matrix $\bf{X}$. We can classify all the samples in $\bf{X}$ by
\begin{eqnarray} \label{eq:class_X}
\begin{split}
{\bf y} = \mbox{sgn} \left( \mathcal{F}^{-1}({\bf \hat{x}}^* \circ {\bf \hat{w}}) + b \right),
\end{split}
\end{eqnarray}
where $\mathcal{F}^{-1}(\cdot)$ denotes the inverse discrete Fourier transform (IDFT), and ${\bf \hat{x}}^*$ denotes the complex conjugate of ${\bf \hat{x}}$.
Given the circulant matrix $\bf{X}$ generated by an $n \times n$ image ${\bf x}$,
the computational complexity of classifying every ${\bf x}_i$ by (\ref{eq:class_x}) is $O(n^4)$,
while that of classifying all samples of $\bf{X}$ by (\ref{eq:class_X}) is $O(n^2 \log n)$.

Given the training set of a circulant matrix ${\bf X} = [{\bf x}_1; {\bf x}_2; \ldots; {\bf x}_{n^2}]$
with the corresponding class labels ${\bf y} = [y_1, y_2, \ldots, y_{n^2}]^\top$, we
use the squared hinge loss and define the SVM model \cite{lee2013study} as follows:
\begin{eqnarray} \label{eq:svm_x}
\begin{split}
\nonumber & \min_{{\bf w}, b} \| {\bf w} \|^2 + C \sum_i \xi_i^2 \\
& \mbox{s.t. } y_i({\bf w}^\top {\bf x}_i + b ) \geq  1 - \xi_i, \mbox{ } \forall i\\
\end{split}
\end{eqnarray}
where ${\boldsymbol{\xi}} = [{\xi}_1, {\xi}_2, \ldots, {\xi}_i \ldots, {\xi}_{n^2}]$
is the vector of slack variables.

Based on the circulant property of ${\bf X}$, the SVM model can be equivalently formulated as:
\begin{eqnarray} \label{eq:svm_X}
\begin{split}
& \min_{{\bf w}, b} \| {\bf w} \|^2 + C \| \boldsymbol{\xi} \|_2^2 \\
& \mbox{s.t. } {\bf y} \circ (\mathcal{F}^{-1} ({\bf \hat{x}^*} \circ {\bf \hat{w}}) + b{\bf {1}} )) \geq  {\bf {1}} - \boldsymbol{\xi}.
\end{split}
\end{eqnarray}
where $\circ$ denotes the element-wise multiplication, and ${\bf {1}}$ denotes a vector of $1$s.

{\flushleft {\textbf{Class labels of the translated images.}}}
Let ${\bf p}^*$ denote the centre position of the object of interest ${\bf x}^*$, and ${\bf p}_i$ as
the position of the translated image ${\bf x}_i$.
In object detection \cite{girshick2014rich, uijlings2013selective},
the overlap function $s({\bf p}^*, {\bf p}_i)$ is used to measure the similarity between
${\bf x}^*$ and ${\bf x}_i$.
Specifically, the positive samples are defined by all ground truth object windows and
the negative samples are defined by those with $s({\bf p}^*, {\bf p}_i)$ below a lower overlap threshold.
In the proposed discriminative tracking model, we need to set upper and lower thresholds of $s({\bf p}^*, {\bf p}_i)$ for assigning binary labels.
In Section~\ref{experiments}, we determine the
optimal upper and lower thresholds for SCF, MSCF and KSCF respectively with experiments.
%

In this work,
we use the following confidence map of object position \cite{zhang2014fast} to define the class label:
\begin{eqnarray} \label{eq:confid}
\nonumber m({\bf p}_i) = \gamma \exp \left( - {\alpha} {\|{\bf p}_i - {\bf p}^*\|^{\beta}} \right),
\end{eqnarray}
where $\gamma$ is a normalization constant, $\alpha$ and $\beta$ are the scale and shape parameters, respectively.
With the confidence map, we define the class labels as follows:
\begin{eqnarray} \label{eq:label}
y_i = {\left\{\begin{matrix}
 1, & \mbox{if } m(p_i, p^*) \geq \theta_u \\
 -1, & \mbox{if } m(p_i, p^*) \leq \theta_l \\
 0,  & \mbox{otherwise}
\end{matrix}\right.},
\end{eqnarray}
where $\theta_l$ and $\theta_u$ are the lower and upper thresholds, respectively.
With this formulation, we can use the circulant matrix
formed by all samples to improve training efficiency,
and discard any samples that are not labeled.

\begin{figure}[t]
\begin{center}
\includegraphics[width=0.5\textwidth]{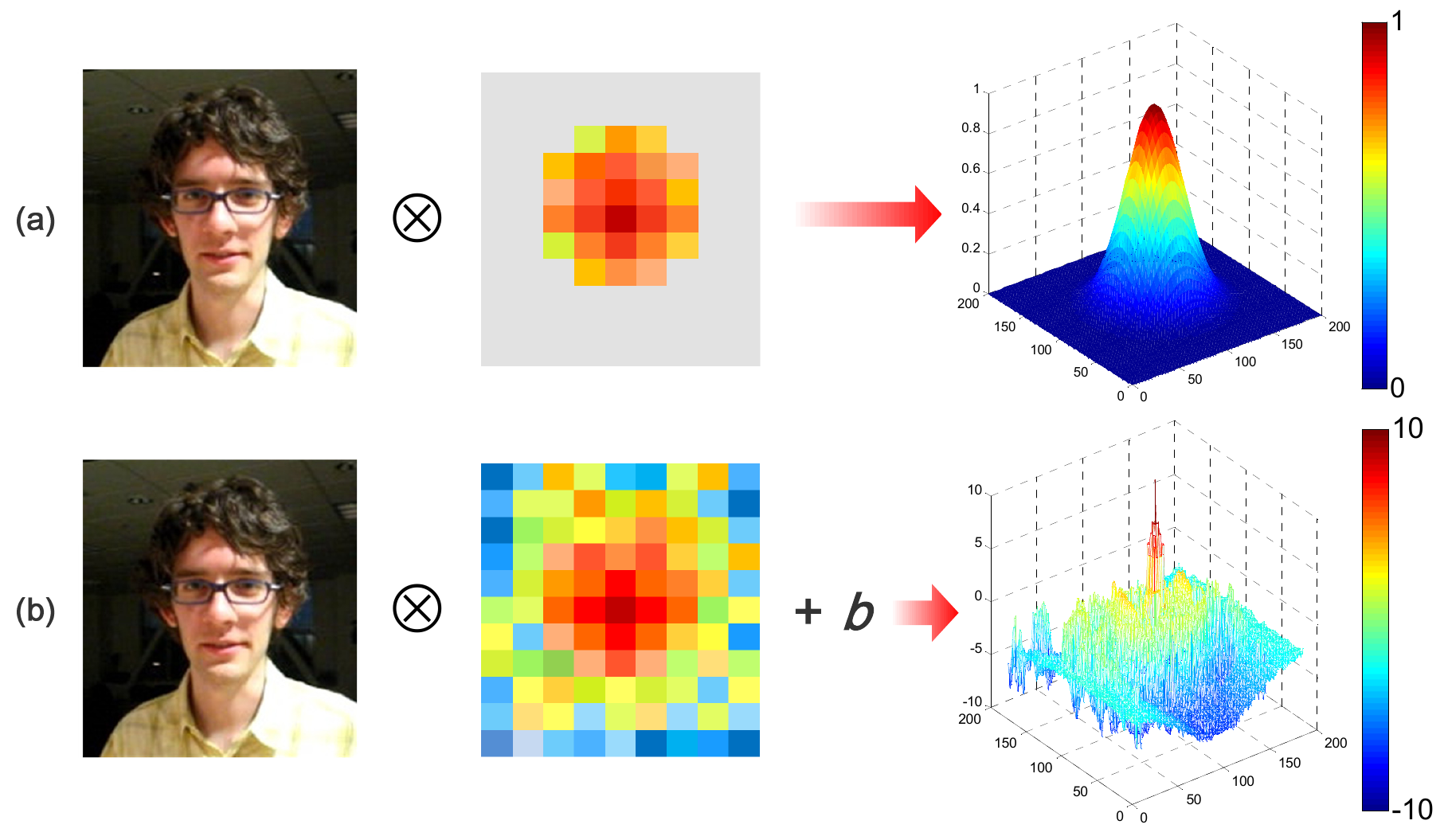}
\end{center}
\caption{Differences between the proposed SCF model and existing CF approaches
\cite{bolme2010visual, henriques2012exploiting, zhang2014fast}.
(a) Existing CF-based models are designed to
learn correlation filters that make the actual output being close to the predefined confidence maps.
(b) The SCF model aims to learn a support correlation filter together with the bias $b$ for distinguishing a target object from the background based on the max margin principle.
%
The peak value in the right response map of (b) locates the target object well.}
\label{fig:modelcf}
\end{figure}

{\flushleft {\textbf{Comparisons with existing CF-based trackers.}}}
As illustrated in Fig.~\ref{fig:modelcf}(a),
existing CF-based trackers generally follow the ridge regression models.
That is, with the continuous confidence map ${\bf m}$, CF-based trackers seek the optimal correlation filter by minimizing the mean squared error (MSE) between the predefined confidence map and actual output,
\begin{equation}
\min_{{\bf w}} \lambda \| {\bf {w}} \|^2 +  \| {\bf Xw} - {\bf m}\|_2^2,
\label{eq:mosse}
\end{equation}
which has the closed form solution,
\begin{eqnarray}
{\bf \hat{w}} =  \frac{ {\bf \hat{x}^*} \circ {\bf \hat{m}} }{ {\bf
    \hat{x}^*} \circ {\bf \hat{x}} + \lambda }.
\label{eq:mosse_fiter}
\end{eqnarray}

As shown in Fig.~\ref{fig:modelcf}(b),
%
the proposed model aims to learn a max-margin SVM classifier $\{ {\bf w}, b \}$
to distinguish the object of interest from the background.
Using the label assignment scheme in (\ref{eq:label}), we can discard uncertain samples in training to alleviate the label ambiguity problem.
The importance of SVM and label ambiguity issues have been demonstrated
in object detection \cite{girshick2014rich}.
%
The proposed model copes with both issues (classification and label ambiguity)
for  effective visual tracking.

\subsection{Alternating optimization}
In this section, we reformulate the model in (\ref{eq:class_X})
and propose an alternating optimization algorithm to learn SCFs efficiently.
To exploit the property of the circulant matrix for learning SCFs,
we let $\boldsymbol{\xi} = {\bf e} + {\bf 1} - {\bf y} \circ (\mathcal{F}^{-1} ({\bf \hat{x}^*} \circ {\bf \hat{w}}) +  b{\bf 1}) $, and the SVM model in  (\ref{eq:svm_X}) is then reformulated as:
\begin{eqnarray}
\label{eq:svm_reform}
\begin{split}
&\min_{{\bf w}, b, {\bf e}} \| {\bf w} \|^2 + C || \left( {\bf y} \circ (\mathcal{F}^{-1} ({\bf \hat{x}^*} \circ {\bf \hat{w}}) +  b{\bf 1}) - {\bf 1}  -  {\bf e} \right) ||^2 \\
&\mbox{s.t. } {\bf e} \geq  0.
\end{split}
\end{eqnarray}
\noindent
With this formulation,
the subproblem on ${\bf e}$ has a closed form solution when $\{ {\bf w}, b \}$ is known, and the subproblem on $\{ {\bf w}, b \}$ has a closed form solution when ${\bf e}$ is known.
Thus the above model can be efficiently solved using the alternating optimization algorithm
by iterating between the following two steps:

{\flushleft {\textbf{Updating ${\bf e}$.}}}
Given $\{ {\bf w}, b \}$, we let ${\bf e}_0 = {\bf y} \circ (\mathcal{F}^{-1}({\bf \hat{x}^*} \circ {\bf \hat{w}}) +  b{\bf 1}) - {\bf 1}$, and the subproblem on ${\bf e}$ becomes:
\begin{eqnarray} \label{eq:prob_e}
\nonumber \min_{\bf e} \|{\bf e} - {\bf e}_0  \|^2, \mbox{s.t. } {\bf e} \geq  0.
\end{eqnarray}
The ${\bf e}$ subproblem has the closed form solution:
\begin{eqnarray} \label{eq:solve_e}
{\bf e} = \max \{ {\bf e}_0, 0 \}.
\end{eqnarray}

{\flushleft {\textbf{Updating $\{ {\bf w}, b \}$.}}}
Given ${\bf e}$, we let ${\bf q} = {\bf y} + {\bf y} \circ {\bf e} $, and the subproblem on $\{ {\bf w}, b \}$ becomes:
\begin{eqnarray} \label{eq:prob_w}
\nonumber \min_{{\bf w}, b} \| {\bf w} \|^2 + C || \left(  \mathcal{F}^{-1} ({\bf \hat{x}^*} \circ {\bf \hat{w}})  + b{\bf 1} - {\bf q} \right) ||^2.
\end{eqnarray}
The subproblem with $\{ {\bf w}, b \}$ is a quadratic programming problem.
One feasible solution is to let ${\bf u} = [{\bf w}; b]$ and derive the closed form solution on ${\bf u}$.
However,  this approach fails to exploit the circulant property of ${\bf X}$.
Thus, we obtain $\{ {\bf w}, b \}$ by solving the following system of equations:
\begin{align}
\label{eq:update_w}
 {\bf \hat{w}} & = \frac {\hat{\bf x}^* \circ (\hat{\bf q} -
 b{\bf \hat{\bf 1}})}{\hat{\bf x}^* \circ \hat{\bf x} + {\bf 1}/C}, \\
\label{eq:update_b0}
b & = {\bf 1}^\top ({\bf q} - \mathcal{F}^{-1} ({\bf \hat{x}^*} \circ {\bf \hat{w}} )),
\end{align}
where ${\bf q} = {\bf y} + {\bf y} \circ {\bf e}$.
Combining the two equations above and with the property of DFT, we have
\begin{eqnarray} \label{eq:update_b}
b = {\bar{ q}},
\end{eqnarray}
where $\bar{q}$ is the mean of ${\bf q}$.
Given $b$, we use (\ref{eq:update_w}) to obtain the closed form solution to ${\bf w}$.

\begin{algorithm}[t]
\caption{SCF model training}
\label{Algm:train_model}

\begin{algorithmic}[1]
 \Require Training image patch ${{\bf x}_{t}}\ \left( n \times n \right)$
class labels ${{\bf y}_{t}}\ \left( n \times n \right)$
 \Ensure
 $\left( \mathbf{\hat{w}},b \right)$.
 \State Initialize ${{\mathbf{\hat{w}}}_{0}}$, ${{b}_{0}}$, $k = 1$.
 \While{not converged}
  \State // Lines 4-5 : updating $\mathbf{e}_{k}$.
  \State $\mathbf{d}={{\bf y}_{t}}\circ \left( {\mathcal{F}^{-1}}\left( {{\bf {\hat{x}^*}}_{t}} \circ {{{\mathbf{\hat{w}}}}_{k-1}}  \right)+b{\bf 1} \right)- {\bf 1}$,
  \State ${{\mathbf{e}}_{k}}=\max \left( 0,\mathbf{d}\right)$, 
  \State // Lines 7-9 : updating $\mathbf{q}_{k}$, $b_{k}$, $\mathbf{p}_{k}$.
  \State ${{\mathbf{q}}_{k}}={{\bf y}_{t}}+{{\bf y}_{t}}\circ {{\mathbf{e}}_{k}}$,
  \State ${{b}_{k}}=mean ({{\mathbf{q}}_{k}})$,
  \State ${{\mathbf{p}}_{k}}={{\mathbf{q}}_{k}}-{{b}_{k}}\mathbf{1}$,
  \State // Line 11 : updating $\mathbf{w}_{k}$.
  \State ${{\mathbf{\hat{w}}}_{k}}=\frac{{{{\hat{\bf x}^*}}_{t}}\circ {{{\mathbf{\hat{p}}}}_{k}}}{{{{\hat{\bf x}^*}}_{t}}\circ {{{\hat{\bf x}}}_{t}}+{{\bf 1}/C}}$.
  \State $k \leftarrow k + 1$
 \EndWhile
\end{algorithmic}
\end{algorithm}

As illustrated in Fig.~\ref{fig:scf_learn}, when the $t$-th frame ${\bf x}^t$ with class labels ${\bf y}^t$ arrives, the proposed algorithm learns support correlation filters by iterating between updating ${\bf e}$
and updating $\{ {\bf w}, b \}$ until convergence.
Given $\{{\bf x}^t,~{\bf y}^t,~{\bf w},~b\}$, the update of ${\bf e}$ can be computed element-wise,
which has the complexity of $O(n^2)$.
Given $\{{\bf x}^t,~{\bf y}^t,~{\bf e}\}$, the complexity of updating $b$ is $O(n^2)$ and that of
updating ${\bf w}$ is $O(n^2 \log n)$.
Thus, the complexity is $O(n^2 \log n)$ per iteration which makes our algorithm
efficient in learning support correlation filters.
The main steps of the proposed learning algorithm for support correlation filters are summarized in Algorithm~\ref{Algm:train_model}.

{\flushleft {\textbf{Convergence.}}}
The proposed algorithm converges to the global optimum with the $q$-linear convergence rate.
For presentation clarity, we give the detailed analysis and proof on its optimality condition, global convergence, and convergence rate in Appendix~\ref{Appendix_A}.
Based on the optimality condition, we define
\begin{eqnarray} \label{eq:res3}
\nonumber \begin{cases}
{\bf {r}}_1 \doteq {\bf w} + C \mathcal{F}^{-1} ( {\bf \hat{x}} \circ {\bf \hat{x}}^{*} \circ {\bf \hat{w}} - {\bf \hat{r}}), \!&\!   \\
{\bf {r}}_2(i) \doteq e_i \!+\! 1 \!-\! ({\bf {y}} \circ (\mathcal{F}^{-1} ( {\bf \hat{x}}^{*} \circ {\bf \hat{w}} ) + b))_i, \!&\! \text{ if } e_i > 0  \\
{\bf {r}}_3(i) \doteq ({\bf {y}} \circ ( \mathcal{F}^{-1} ( {\bf \hat{x}}^{*} \circ {\bf \hat{w}} ) + b))_i - 1, \!&\! \text{ if } e_i =  0
\end{cases}
\end{eqnarray}
and adopt the following stopping criterion:
\begin{eqnarray}
\label{eq:msvm_X2}
\begin{split}
\max\{\| {\bf {r}}_1 \|_{\infty}, \max_{e_i > 0}\{\| {\bf {r}}_1 (i) \|\},  \max_{e_i = 0}\{{\bf {r}}_3 (i)\} \} \leq \epsilon.
\end{split}
\end{eqnarray}
{\flushleft {\textbf{Comparisons with MMCF \cite{rodriguez2013maximum}.}}}
The proposed SCF model and learning algorithm are different from the MMCF approach
in three aspects.
%
First,  the training samples for MMCF are $N$ images of $n \times n$ pixels,
while those for SCF are $n^2$ translated images of $n \times n$ pixels.
We exploit the circulant property of the data matrix ${\bf X}$ to develop an efficient learning algorithm.
Second, we propose an alternating optimization algorithm to solve the proposed model, which has the complexity of $O(n^2 \log n)$.
In contrast, the MMCF method adopts the conventional SMO algorithm with
the complexity of $O(N^2d)$ where $d$ is the dimension of the sample.
For visual tracking considered in this work, we have $N = n^2$ and $d = n^2$, and the complexity of MMCF is $O(n^6)$, which is computationally expensive for real-time applications.
Third, the proposed model has the squared hinge loss and regularizer terms,
while the MMCF method adopts the hinge loss and includes an extra average correlation energy term.
%
\subsection{Multi-channel SCF}
Different local descriptors, e.g., color attributes, HOG, and SIFT \cite{dalal2005histograms,danelljan2014adaptive,zhou2009object}, provide rich
image features for effective visual tracking.
We treat local descriptors as multi-channel images where multiple measurements
are associated to each pixel.
To exploit multi-dimensional features, we propose the multi-channel SCF
as follows:
\begin{eqnarray} \label{eq:msvm_X}
\begin{split}
&\min_{{\bf w}, b} \| {\bf w} \|^2 + C \| \boldsymbol{\xi} \|_2^2\\
 & \mbox{s.t. } {\bf y} \circ ( \mathcal{F}^{-1} (\sum_{l=1}^{L} {\ ({\bf \hat{x}}^l)}^* \circ {\bf \hat{w}}^l) + b{\bf 1}) \geq {\bf 1} - \boldsymbol{\xi}
\end{split}
\end{eqnarray}
where $L$ is the number of channels, and ${\bf {x}}^{l}$ and ${\bf {w}}^{l}$ denote the $l$-th channel of the image and correlation filter, respectively.
To learn the proposed MSCF model,
we adopt the same equations on updating $\bf e$ and $b$,
and compute ${\bf w}$ by solving the following problem:
\begin{eqnarray} \label{eq:mprob_w}
\nonumber \min_{{\bf w}} \sum_{l=1}^L \| {\bf \hat{w}}^l \|^2 + C \| \sum_{l=1}^L {({{\bf \hat{x}}^l})}^* \circ {\bf \hat{w}}^l  - {\bf \hat{r}} \|^2,
\end{eqnarray}
where ${\bf \hat{w}} = [{\bf \hat{w}}^1; {\bf \hat{w}}^2, \ldots, {\bf \hat{w}}^L]$, and ${\bf \hat{r}} = {\bf \hat{q}} - b {\bf \hat{1}}$.

Let ${\bf \hat{X}} = [\mbox{Diag}({\bf \hat{x}}^1) \mbox{ } \mbox{Diag}({\bf \hat{x}}^2) \mbox{ } \ldots \mbox{ } \mbox{Diag}({\bf \hat{x}}^L)]$.
The closed form solution for ${\bf \hat{w}}$ can be directly obtained by
\begin{eqnarray} \label{eq:mupdate_w0}
{\bf \hat{w}} = ({\bf \hat{X}}^H {\bf \hat{X}} +\frac{1}{C} {\bf I})^{-1} {\bf \hat{X}}^H {\bf \hat{r}}.
\end{eqnarray}
\noindent
where $ {\bf I}$ is the identity matrix.
Note that ${\bf \hat{X}}$ is an $n^2 \times Ln^2$ matrix.
It is not practical to compute the inverse of ${\bf \hat{X}}^H {\bf \hat{X}}$ to update ${\bf \hat{w}}$.
In the multi-channel correlation filters,
it is noted that ${\bf \hat{X}}$ has the diagonal block structure,
and the $j$-th element of ${\bf \hat{r}}$ depends only
on ${\bf \hat{w}}(j) = [{\bf \hat{w}}^1(j); {\bf \hat{w}}^2(j); \ldots; {\bf \hat{w}}^L(j)]$
and ${\bf \hat{x}}(j) = [{\bf \hat{x}}^1(j); {\bf \hat{x}}^2(j); \ldots; {\bf \hat{x}}^L(j)]$.
Thus, the subproblem on ${\bf \hat{w}}$ can be further decomposed into $n^2$ systems of equations:
\begin{eqnarray} \label{eq:mprob_w_sub}
\left( {\bf \hat{x}}(j) {\bf \hat{x}}(j)^H +\frac{1}{C} {\bf I} \right) {\bf \hat{w}}(j) = {\bf \hat{x}}(j) {\bf \hat{r}}(j).
\end{eqnarray}

In \cite{galoogahi2013multi}, Galoogahi \textit{et al.} solve these $n^2$ systems of equations by an algorithm with the complexity of $O(n^2L^3 + Ln^2 \log n)$.
We note that the matrix on the left hand of (\ref{eq:mprob_w_sub}) is a rank-one matrix and a scaled identity matrix.
Based on the Sherman-Morrison formula \cite{petersen2008matrix}, we have
\begin{eqnarray} \label{eq:sherman}
\nonumber \left( {\bf \hat{x}}(j) {\bf \hat{x}}(j)^H +\frac{1}{C} {\bf I} \right)^{-1} \!=\! C \left( {\bf I} \!-\! \frac{C{\bf \hat{x}}(j) {\bf \hat{x}}(j)^H}{1+C {\bf \hat{x}}(j)^H {\bf \hat{x}}(j)} \right).
\end{eqnarray}
\noindent
The closed form solution for ${\bf \hat{w}}(j)$ is then obtained by
\begin{eqnarray} \label{eq:mprob_w_sol}
{\bf \hat{w}}(j) = \frac{C{\bf \hat{x}}(j) {\bf \hat{r}}(j)}{1+C {\bf \hat{x}}(j)^H {\bf \hat{x}}(j)}.
\end{eqnarray}
\noindent

It should be noted that all ${\bf \hat{x}}^l$s can be pre-computed with the complexity of $O(n^2 \log n)$.
As such, the proposed
algorithm only involves one DFT, one IDFT and several element-wise operations per iteration, and the complexity is $O(n^2 \log n)$.
\begin{table*}[thp] \small
\renewcommand{\arraystretch}{1}
\caption{Results of MSCF and DCF with different feature representations.}
\label{table_MSCF_feature} \centering
\begin{tabular}{c|c|c|c|c|c|c|c|c}
\hline
\multicolumn{1}{c|}{  Algorithms} & \multicolumn{4}{|c|}{  MSCF} & \multicolumn{4}{|c}{  DCF \cite{henriques2015high}} \\
\hline
  Features &   Raw pixels &   CN  &   HOG  &   HOG $+$ CN  &   Raw pixels &   CN  &   HOG  &   HOG $+$ CN   \\
\hline\hline
  Mean DP (\%)   & 64.9  & 66.3  & 78.4  & \textcolor[rgb]{1.00,0.00,0.00}{{\bf 80.6}}  & 44.4  & 48.0  & 71.9  & 76.2\\
\hline
  Mean AUC (\%)       & 44.6  & 44.9  & 53.7  & \textcolor[rgb]{1.00,0.00,0.00}{{\bf 55.5}}  & 31.2  & 34.8  & 50.1  & 53.2\\
\hline
  Mean FPS (\it s)            & 76    & 62    & 64    & 54          & 278   & 210   & \textcolor[rgb]{1.00,0.00,0.00}{{\bf 292}}  & 151\\
\hline
\end{tabular}
\end{table*}
\begin{table*}[thp] \small
\renewcommand{\arraystretch}{1}
\caption{ Results of KSCF and KCF with different feature representations.}
\label{table_KSCF_feature} \centering
\begin{tabular}{c|c|c|c|c|c|c|c|c}
\hline
\multicolumn{1}{c|}{  Algorithms} & \multicolumn{4}{|c|}{  KSCF} & \multicolumn{4}{|c}{  KCF \cite{henriques2015high}} \\
\hline
  Features &   Raw pixels &   CN   &   HOG  &   HOG $+$ CN   &   Raw pixels &   CN   &   HOG  &   HOG $+$ CN  \\
\hline\hline
  Mean DP (\%)   & 64.4  & 68.1  & 79.3  & \textcolor[rgb]{1.00,0.00,0.00}{{\bf 85.0}}      & 55.3  & 57.3  & 73.2  & 75.8 \\
\hline
  Mean AUC (\%)       & 45.3  & 46.9  & 53.2  & \textcolor[rgb]{1.00,0.00,0.00}{{\bf 57.5}}      & 40.0  &  41.8 & 50.7  & 53.0 \\
\hline
  Mean FPS (\it s)            & 40    & 37     &  44   & 35              & 154   & 120   & \textcolor[rgb]{1.00,0.00,0.00}{{\bf 172}}  & 102 \\
\hline
\end{tabular}
\end{table*}
\subsection{Kernelized SCF}
Given the kernel function $K({\bf x}, {\bf x}^{\prime}) = \langle \psi({\bf x}), \psi({\bf x}^{\prime}) \rangle$,
the proposed kernelized SCF model can be extended to learn the nonlinear decision function:
%
\begin{eqnarray} \label{eq:nclass_x}
\nonumber f({\bf x}) = {\bf w}^\top \psi({\bf x}) + b = \sum_i {\alpha_i K({\bf x}, {\bf x}_i)} + b,
\end{eqnarray}
where $\psi({\bf x})$ stands for the nonlinear feature mapping implicitly determined by the kernel function $K({\bf x}, {\bf x}^{\prime})$,
and $\boldsymbol{\alpha} = [\alpha_1, \alpha_2,  \ldots, \alpha_{n^2}]^\top$ is the coefficient vector to be learned.

Denote by ${\bf K}$ the kernel matrix with $K_{ij} = K({\bf x}_i, {\bf x}_j)$.
As noted in \cite{henriques2012exploiting}, for some kernel functions (e.g., Gaussian RBF and polynomial) which are permutation invariant, the kernel matrix ${\bf K}$ is circulant.
Let ${\bf k}^{\bf xx}$ be the first row of the circulant matrix ${\bf K}$.
Therefore, the matrix-vector multiplication ${\bf K \boldsymbol{\alpha}}$
can be efficiently computed via DFT:
\begin{eqnarray} \label{eq:Kalpha}
{\bf K} \boldsymbol{\alpha} = \mathcal{F}^{-1}({\bf \hat{k}}^{\bf xx} \circ {\hat{\boldsymbol{\alpha}}}),
\end{eqnarray}
and we have,
\begin{eqnarray} \label{eq:hilbertnorm}
\| {\bf w} \|^2 = \boldsymbol{\alpha}^\top {\bf K} \boldsymbol{\alpha} = \boldsymbol{\alpha}^\top \mathcal{F}^{-1}({\bf \hat{k}}^{\bf xx} \circ {\bf \hat{ \boldsymbol{\alpha} }}).
\end{eqnarray}
\noindent
Based on (\ref{eq:Kalpha}) and (\ref{eq:hilbertnorm}), the proposed kernelized SCF
model is formulated as
\begin{eqnarray}
\label{eq:msvm_X3}
\begin{split}
\min_{{\boldsymbol{\alpha}}, b} \mbox{ }& {\boldsymbol{\alpha}}^\top \mathcal{F}^{-1}({\bf \hat{k}}^{\bf xx} \circ {\hat{\boldsymbol{\alpha}}})\\
 &+ C ({\bf y} \circ ( \mathcal{F}^{-1} ({\bf \hat{k}}^{\bf xx} \circ {\hat{\boldsymbol{\alpha}}}) + b{\bf 1} ) - {\bf 1} - {\bf e})^2\\
 &\mbox{s.t. } {\bf e} \geq 0.
\end{split}
\end{eqnarray}

To learn KSCF, we use the alternating optimization method by
iteratively solving ${\bf e}$ and $\{ {\bf w}, b \}$.
The solution of the subproblem with ${\bf e}$ is similar to that in the SCF model,
and we update $b$ and ${\bf w}$ using the closed form solution of kernel ridge regression.
Based on the representation theorem \cite{rifkin2003regularized},
the optimal solution ${\bf w}$ in the kernel space
can be expressed as the linear combination of the feature maps of the samples:
${\bf {w}^{*}}=\sum\limits_{i}{\alpha _{i}^{*}\varphi \left( {{x}_{i}} \right)}$.
Namely, only the coefficient vector $\boldsymbol{\alpha}$ needs to be learned.
In \cite{rifkin2003regularized}, the solution to the kernelized ridge regression in the dual space is
given by
\begin{eqnarray} \label{eq:alpha_closeform}
\nonumber {\boldsymbol{\alpha}} ={{\left( {\bf K}\ +\ \lambda {\bf I} \right)}^{-1}}{\bf y}.
\end{eqnarray}
Thus, the closed form solution to our sub-problem on ${\boldsymbol{\alpha}}$ can be formulated as
\begin{eqnarray} \label{eq:alpha_closeform_w}
\nonumber {\boldsymbol{\alpha}} ={{\left( {\bf K}\ +\ \lambda {\bf I} \right)}^{-1}}({\bf q}-b{\bf 1}),
\end{eqnarray}
where ${\bf q}\ = {\bf y} + {\bf y} \circ {\bf e}$ and {\bf 1} denotes a vector of 1s.
As the kernel matrix {\bf K} is circulant and can be diagonalized, the optimal solution of ${\boldsymbol{\alpha}}$ in the Fourier transform domain can be computed by
\begin{eqnarray} \label{eq:alphaf_closeform}
{{\hat{\boldsymbol{\alpha}}}^{*}}=\frac{{\bf \hat{t}}}{{{\bf \hat{{k}}}^{xx}} + {1/C}},
\end{eqnarray}
where ${\bf \hat{t}}\ = {\bf \hat{y}} + {\bf \hat{y}} \circ {\bf \hat{e}} - b{\bf \hat{1}}$, ${\bf {k}^{xx}}$ is the kernel correlation of ${\bf x}$ with itself in the Fourier domain which is known as the kernel auto-correlation.

For image features with $L$ channels, the complexity to compute kernel matrix is $O(Ln^2 \log n)$.
After that, the learning process only requires element-wise operations,
one DFT and one IDFT per iteration, and the complexity is $O(n^2 \log n)$.
Thus, the proposed KCSF model leverages rich features from the nonlinear filters without
increasing computational load significantly.

Furthermore, to handle large scale changes, we
%
develop the SKSCF model by maintaining a scaling pool in a way similar to
%
the scale-adaptive CF scheme \cite{li2014scale}, and the
bilinear interpolation is used to resize samples across scales.

\section{Performance Evaluation}
\label{experiments}
We use the benchmark dataset and protocols \cite{wu2013online}
to evaluate the proposed SCF algorithms.
First, we evaluate several variants of the proposed method, i.e., SCF, MSCF, KSCF, and SKSCF,
to analyze the effect of feature representations and kernel functions.
Next, comprehensive experiments are conducted to compare
the proposed methods with other CF-based trackers.
Finally, the KSCF and SKSCF algorithms
are compared with existing SVM-based and the state-of-the-art methods.
%
The tracking results can be found at \url{http://faculty.ucmerced.edu/project/scf/} and the source code will be made available to the public.

\subsection{Experimental setup}
{\flushleft {\textbf{Datasets and evaluated tracking methods.}}}
To assess the performance of the proposed methods,
experiments are carried out on a benchmark dataset \cite{wu2013online}
of 50 challenging image sequences annotated with 11 attributes.
For the first frame of each sequence, the bounding box of the target object
is provided for fair comparisons.
For comprehensive comparisons,
we evaluate the baseline SCF, multi-channel SCF, kernelized SCF and SKSCF methods.
The SCF and MSCF methods are designed in the linear space with raw pixels,
and multi-channel features are based on HOG \cite{dalal2005histograms}
as well as color names (CN) \cite{danelljan2014adaptive}, respectively.
The KSCF and SKSCF algorithms are evaluated by using the Gaussian kernel
on multi-channel feature representations.
Furthermore, we compare the proposed trackers with the other trackers based on correlation filters (e.g., MOSSE \cite{bolme2010visual}, CSK \cite{henriques2012exploiting}, KCF \cite{henriques2015high}, DCF \cite{henriques2015high}, STC \cite{zhang2014fast} and CN \cite{danelljan2014adaptive}),
existing SVM based trackers (e.g., Struck \cite{hare2011struck} and MEEM \cite{zhang2014meem}),
and other state-of-the-art methods
(e.g., TGPR \cite{gao2014transfer}, SCM \cite{zhong2012robust}, TLD \cite{kalal2012tracking}, L1APG  \cite{bao2012real}, MIL \cite{babenko2009visual}, ASLA \cite{jia2012visual} and CT \cite{zhang2012real}).

{\flushleft {\textbf{Evaluation protocols.}}}
We use the one-pass evaluation (OPE) protocol \cite{wu2013online}
which reports the precision and success plots
based on the position error and bounding box overlap metrics with
respect to the ground truth object locations.
For precision plots, the distance precision at a threshold of 20 pixels (DP) is reported.
For success plots, the area under the curve (AUC) is computed.
%
In addition, the frames per second (FPS) that each method is able to process is discussed.

{\flushleft {\textbf{Parameter settings.}}} The experiments are carried out on a desktop computer with
an Intel Xenon 2 core 3.30 GHz CPU and 32 GB RAM.
The proposed SCF-based trackers involve a few model parameters, i.e., trade-off parameter $C$, scale parameter $\alpha$ and shape parameter $\beta$ of confidence maps,
and lower and upper thresholds ($\theta_l$, $\theta_u$) in (\ref{eq:label}).
In addition, the KSCF method has one extra parameter $\sigma$ for the
Gaussian RBF kernel function,
%
and SKSCF contains a scaling pool parameter $s$.
For online tracking, the model is updated by linear interpolation with the adaption rate $\rho$
\cite{wu2013online}.

In all experiments, the model parameters are fixed for each SCF-based tracker.
For all SCF-based trackers, the trade-off $C$ and shape parameter $\beta$ are fixed
to ${10^4}$ and $2$, respectively.
The thresholds ($\theta_l$, $\theta_u$) in (\ref{eq:label}) are set to $(0.3, 0.7)$ for SCF, $(0.4, 0.9)$ for MSCF and $(0.5, 0.7)$ for KSCF, SKSCF.
The scale parameter $\alpha$ is set to be $50/(mn)$, which is adaptive to the size $m \times n$
of each target object.
%
The scaling pool $s$ is fixed as $[0.985, 0.990, 0.995, 1, 1.005, 1.010, 1.015]$.
The adaption rate $\rho$ is set to $0.075$ for raw pixel features, and $0.02$ for multi-channel features, respectively.
The kernel parameter $\sigma$ of KSCF is set to $0.2$.
As for HOG parameters, the orientations and cell size are set to 9 and 4.
\subsection{Evaluation on SCF-based trackers}
In this section, we first evaluate the effect of feature representations and kernel functions, and then compare four variants of the SCF-based trackers, i.e., SCF, MSCF, KSCF, and SKSCF, in terms of both accuracy and efficiency.
The results of the corresponding CF-based trackers are also reported for all SCF-based methods.
\begin{table}[t] \small
\renewcommand{\arraystretch}{1}
\caption{Results of KSCF with different kernels.}
\label{table_KSCF_kernels} \centering
\begin{tabular}{c|c|c|c}
\hline
  Kernels &   Linear &   Polynomial  &   Gaussian   \\
\hline\hline
Mean DP (\%)   & 82.0  & 84.2  & \textcolor[rgb]{1.00,0.00,0.00}{{\bf 85.0}}  \\
\hline
Mean AUC (\%)       & 56.2  & 57.1  & \textcolor[rgb]{1.00,0.00,0.00}{{\bf 57.5}}  \\
\hline
Mean FPS (\it s)            & \textcolor[rgb]{1.00,0.00,0.00}{{\bf 94}}    & 55  & 35  \\
\hline
\end{tabular}
\end{table}
\begin{figure}[htbp]
\centering
\begin{tabular}{@{}c@{}c@{}}
    \includegraphics[scale=0.33]{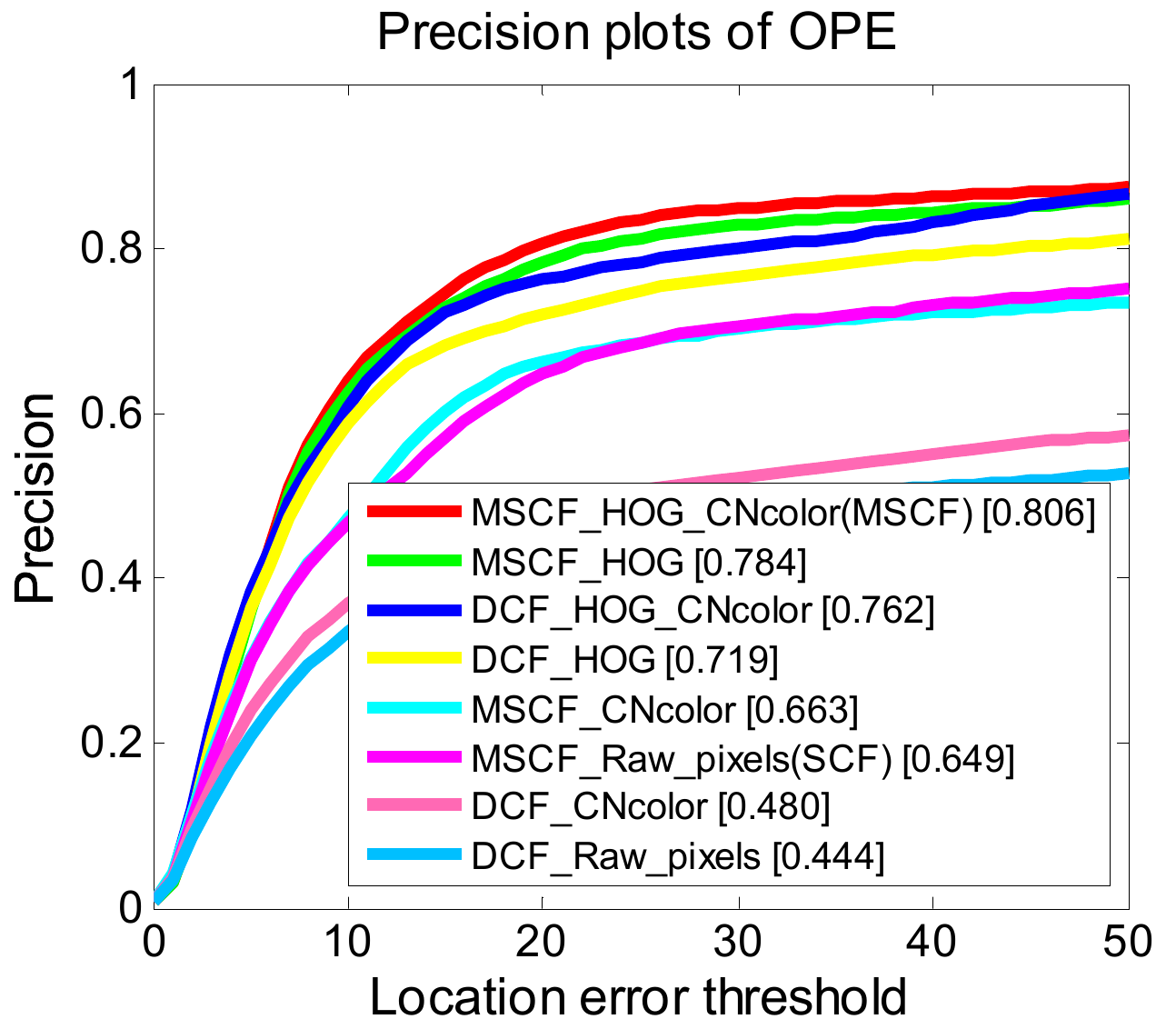} &
    \includegraphics[scale=0.33]{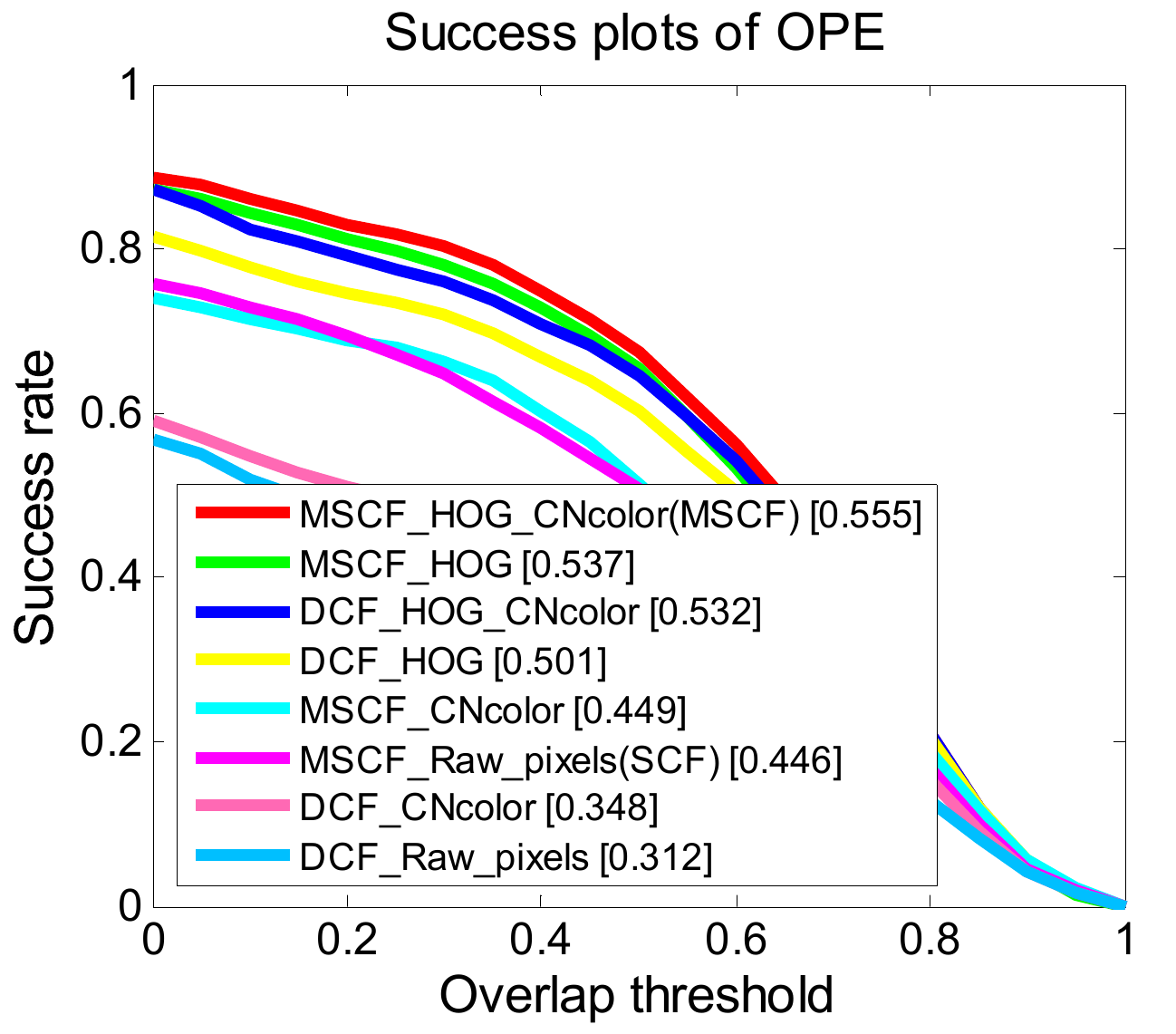}
\end{tabular}
\caption{OPE plots of the MSCF and DCF  \cite{henriques2015high} with different feature representations.
the AUC values are shown next to the legends.}
\label{fig:linear_feature_compare}
\end{figure}

We consider three typical feature representations, i.e., raw pixels, HOG features
\cite{dalal2005histograms}, and color names (CN) \cite{danelljan2014adaptive}.
The results of the MSCF and KSCF methods are listed in Table~\ref{table_MSCF_feature} and Table~\ref{table_KSCF_feature}.
The result for each feature representation is optimal by varying the parameters
$\beta$, $\rho$, $\theta_l$ and $\theta_u$ from $[0.5, 1, 1.5, 2]$, $[0.02, 0.04, 0.075]$, $[0.7, 0.8, 0.9]$
and $[0.3, 0.4, 0.5]$.
These parameters are then fixed for all the following experiments.
For KSCF, the Gaussian RBF kernel with $\sigma \!=\! 0.2$ is adopted.

\begin{figure}[htbp]
\centering
\begin{tabular}{@{}c@{}c@{}}
\includegraphics[scale=0.33]{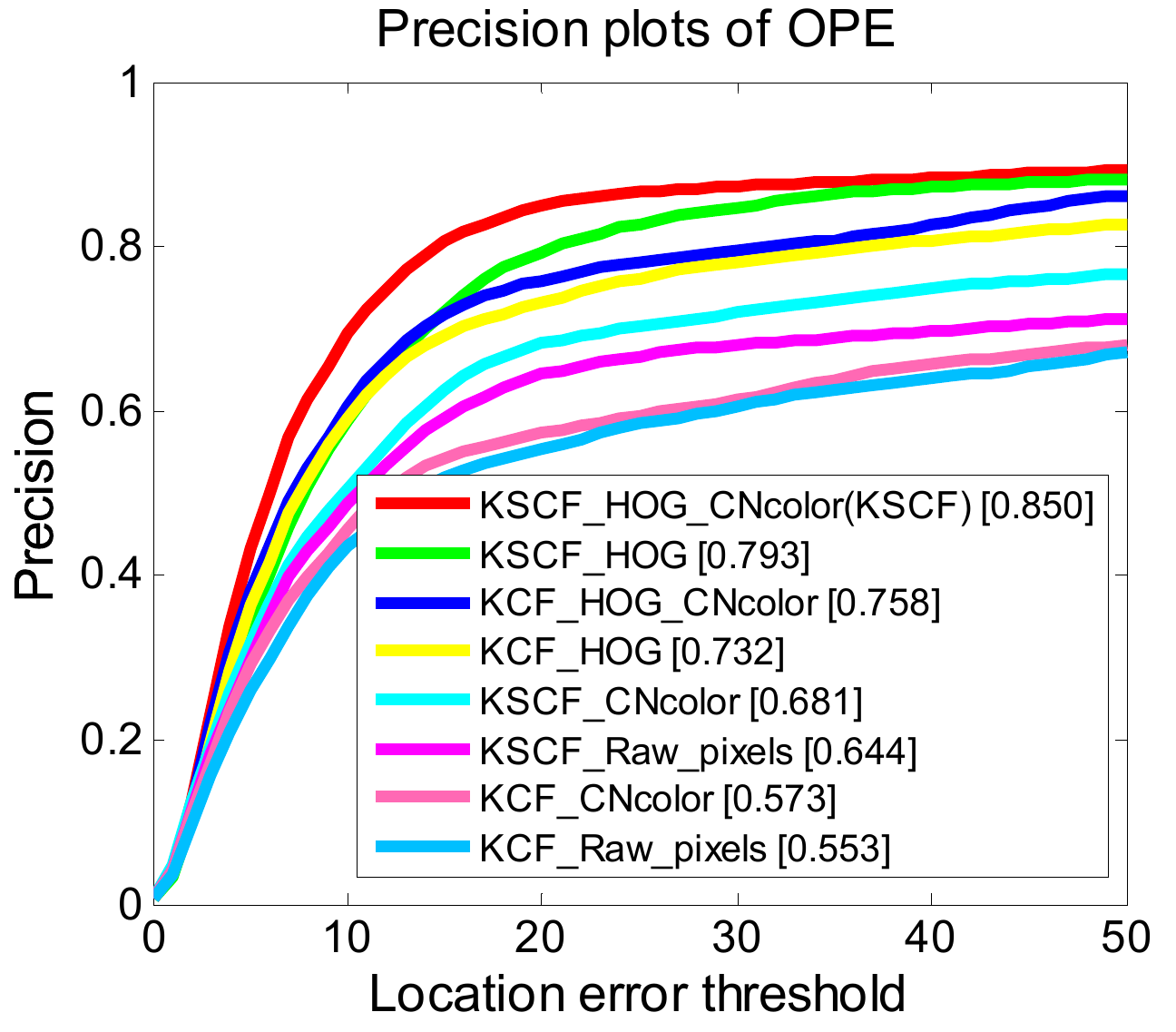} &
\includegraphics[scale=0.33]{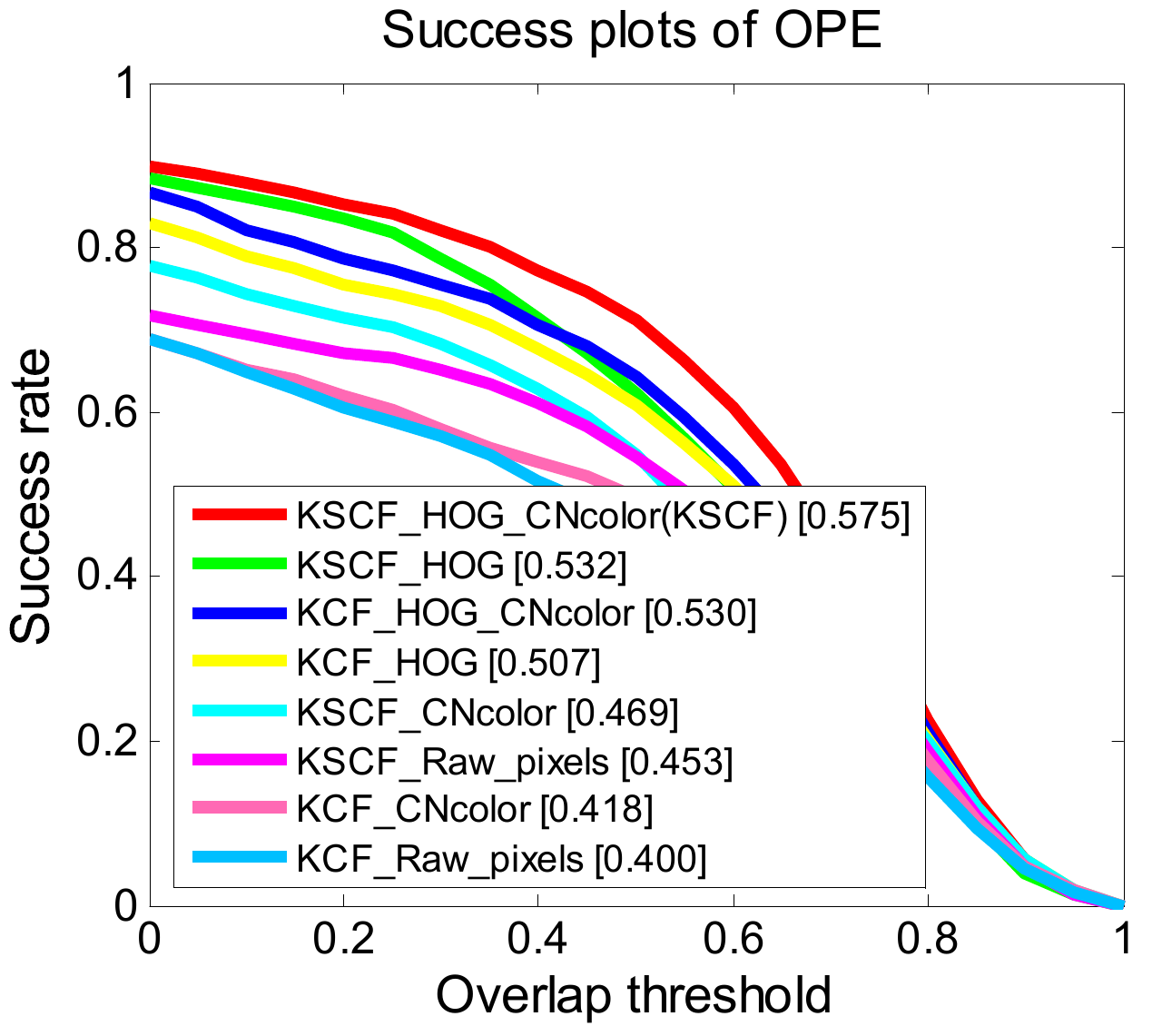}
\end{tabular}
\caption{OPE plots of the KSCF and KCF \cite{henriques2015high} methods
with different feature representations.}
\label{fig:kernel_feature_compare}
\end{figure}

The OPE plots of MSCF with linear DCF \cite{henriques2015high} and KSCF with nonlinear KCF \cite{henriques2015high} are shown in Fig. \ref{fig:linear_feature_compare} and Fig. \ref{fig:kernel_feature_compare}.
Compared with raw pixels and color features, the method with HOG representation significantly improves the tracking performance in terms of mean DP and mean AUC.
For MSCF, the implementation using color names and HOG features outperforms raw pixels
by $1.4\%$ and $13.5\%$ in terms of mean DP.
For KSCF, the tracker using color names and HOG features outperforms raw pixels by
$3.7\%$ and $14.9\%$ in terms of mean DP.

\begin{figure}[htbp]
\centering
\begin{tabular}{@{}c@{}c@{}}
 \includegraphics[scale=0.33]{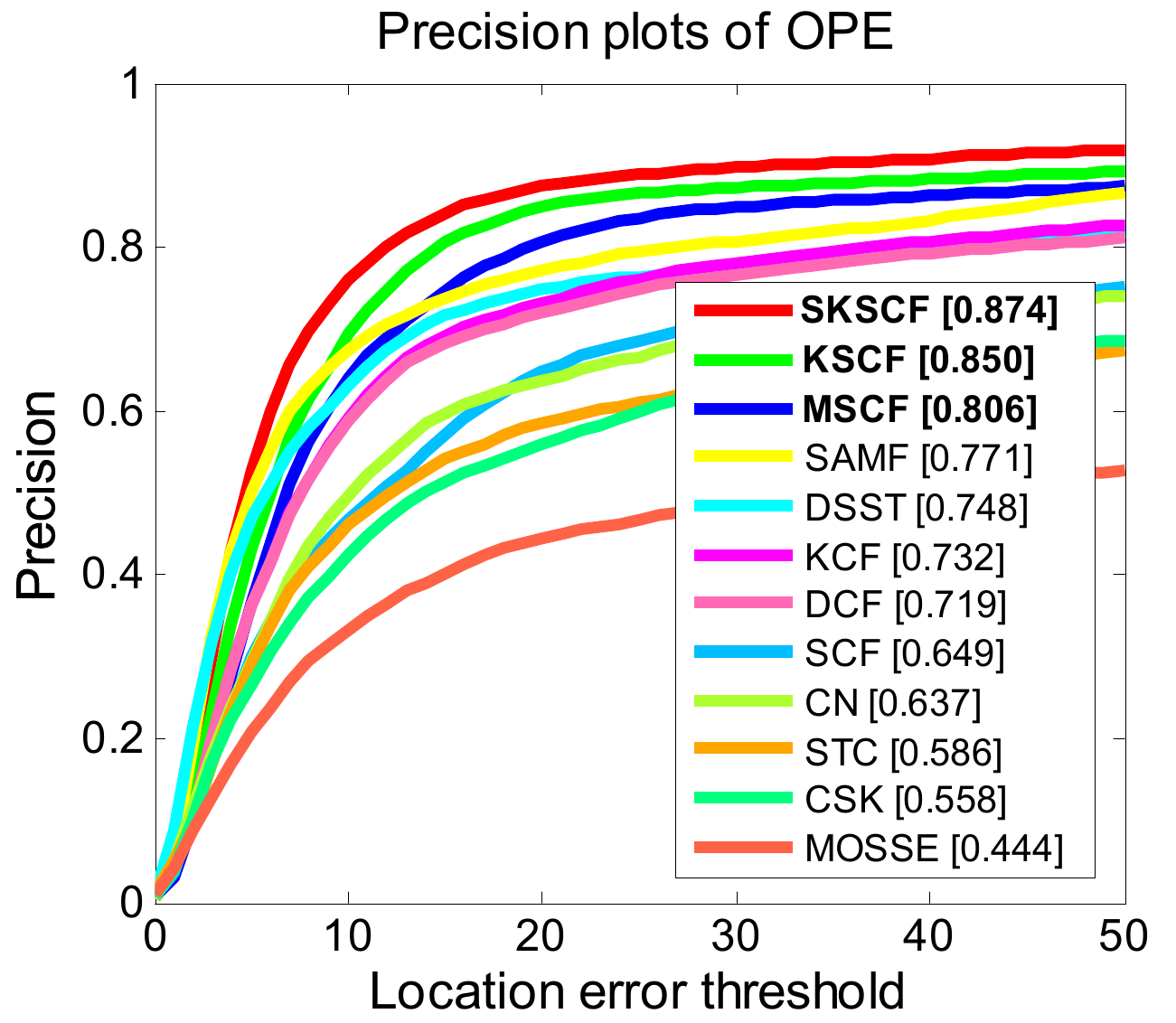} &
 \includegraphics[scale=0.33]{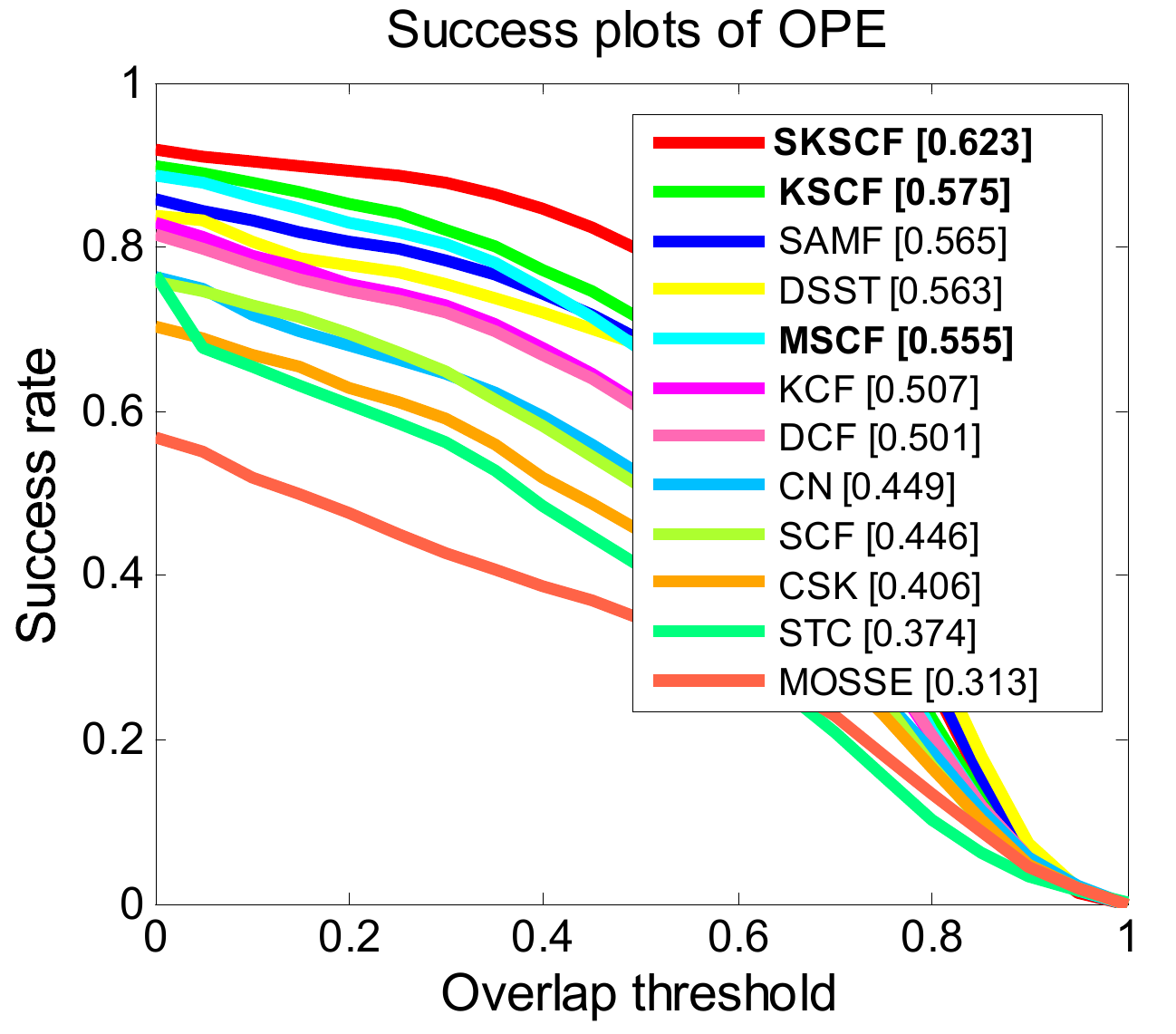}
\end{tabular}
\caption{OPE plots of the SCF methods (i.e., SCF, MSCF, KSCF, and SKSCF)
and other CF-based trackers (i.e., MOSSE \cite{bolme2010visual}, CSK \cite{henriques2012exploiting}, DCF \cite{henriques2015high}, KCF \cite{henriques2015high}, STC \cite{zhang2014fast}, CN \cite{danelljan2014adaptive}, DSST \cite{danelljan2014accurate} and SAMF \cite{li2014scale}).}
\label{fig:plots1}
\end{figure}
%
The MSCF tracker with the combination of color names and HOG is further improved to $80.6\%$ in terms of DP.
Similarly, the performance of the KSCF method is improved to $85.0\%$ in terms of DP with the use of color names and HOG features.
Compared with the DCF \cite{henriques2015high} and KCF \cite{henriques2015high} methods,
the proposed MSCF and KSCF algorithms achieve higher DP and AUC values
for each feature representation.
Table~\ref{table_MSCF_feature} and~\ref{table_KSCF_feature} show
that both KSCF and MSCF methods perform in real-time even using the representation based on
HOG and CN features.
\begin{table*}[t] \small
\renewcommand{\arraystretch}{1}
\caption{Performance of tracking methods based on correlation filters:
Top three results are shown in red, blue and orange.}
\label{table:scf-variants} \centering
\begin{tabular}{c|c|c|c|c|c|c|c|c|c|c|c|c}
\hline
{Algorithms} & \tabincell{c}{{SKSCF}\\{}} & \tabincell{c}{{KSCF}\\{}} & \tabincell{c}{{MSCF}\\{}} & \tabincell{c}{{SCF}\\{}}  & \tabincell{c}{SAMF\\ \cite{li2014scale}} & \tabincell{c}{DSST\\ \cite{danelljan2014accurate}} & \tabincell{c}{KCF\\ \cite{henriques2015high}} & \tabincell{c}{DCF\\ \cite{henriques2015high}} & \tabincell{c}{CN\\ \cite{danelljan2014adaptive}} & \tabincell{c}{STC\\ \cite{zhang2014fast}} & \tabincell{c}{CSK\\ \cite{henriques2012exploiting}}  & \tabincell{c}{MOSSE \\ \cite{bolme2010visual}} \\
\hline \hline
 Mean DP (\%) & \textcolor[rgb]{1.00,0.00,0.00}{\bf 87.4} & \textcolor[rgb]{0.00,0.07,1.00}{\bf 85.0} & \textcolor[rgb]{1.00,0.50,0.00}{\bf 80.6} & 62.8 & 77.1 & 74.8 & 73.2 & 71.9 & 63.7 & 58.6 & 55.8 & 44.4\\
\hline
 Mean AUC (\%) & \textcolor[rgb]{1.00,0.00,0.00}{\bf 62.3} & \textcolor[rgb]{0.00,0.07,1.00}{\bf 57.5} & 55.5 & 48.9  & \textcolor[rgb]{1.00,0.50,0.00}{\bf 56.5} & 56.3 & 50.7 & 50.1 & 44.9 & 37.4 & 40.6 & 31.3 \\
\hline
Mean FPS ({\it s}) & 8 & 35 & 54 & 76 & 14 & 30 & 172 & \textcolor[rgb]{1.00,0.50,0.00}{\bf 292} & 79 & \textcolor[rgb]{1.00,0.00,0.00}{\bf 557} & 151 & \textcolor[rgb]{0.00,0.07,1.00}{\bf 421} \\
\hline
\end{tabular}
\end{table*}
\begin{table}[t] \small
\renewcommand{\arraystretch}{1}
\caption{ Comparison of SVM-based trackers.}
\label{table:svm_based_trackers} \centering
\begin{tabular}{c|c|c|c|c}
\hline
  Algorithms  & \tabincell{c}{{SKSCF}\\{}}  & \tabincell{c}{{KSCF}\\{}}  &   \tabincell{c}{MEEM \\ \cite{zhang2014meem}} &   \tabincell{c}{Struck \\ \cite{hare2011struck}} \\
\hline\hline
  Mean DP (\%)   & \textcolor[rgb]{1.00,0.00,0.00}{{\bf 87.4}}  & 85.0   & 83.3   & 67.4    \\
\hline
  Mean AUC (\%)       & \textcolor[rgb]{1.00,0.00,0.00}{{\bf 62.3}}  & 57.5   & 57.2   & 48.6    \\
\hline
  Mean FPS (\it s)            & 8     & \textcolor[rgb]{1.00,0.00,0.00}{{\bf 35}}     & 10     & 10       \\
\hline
\end{tabular}
\end{table}
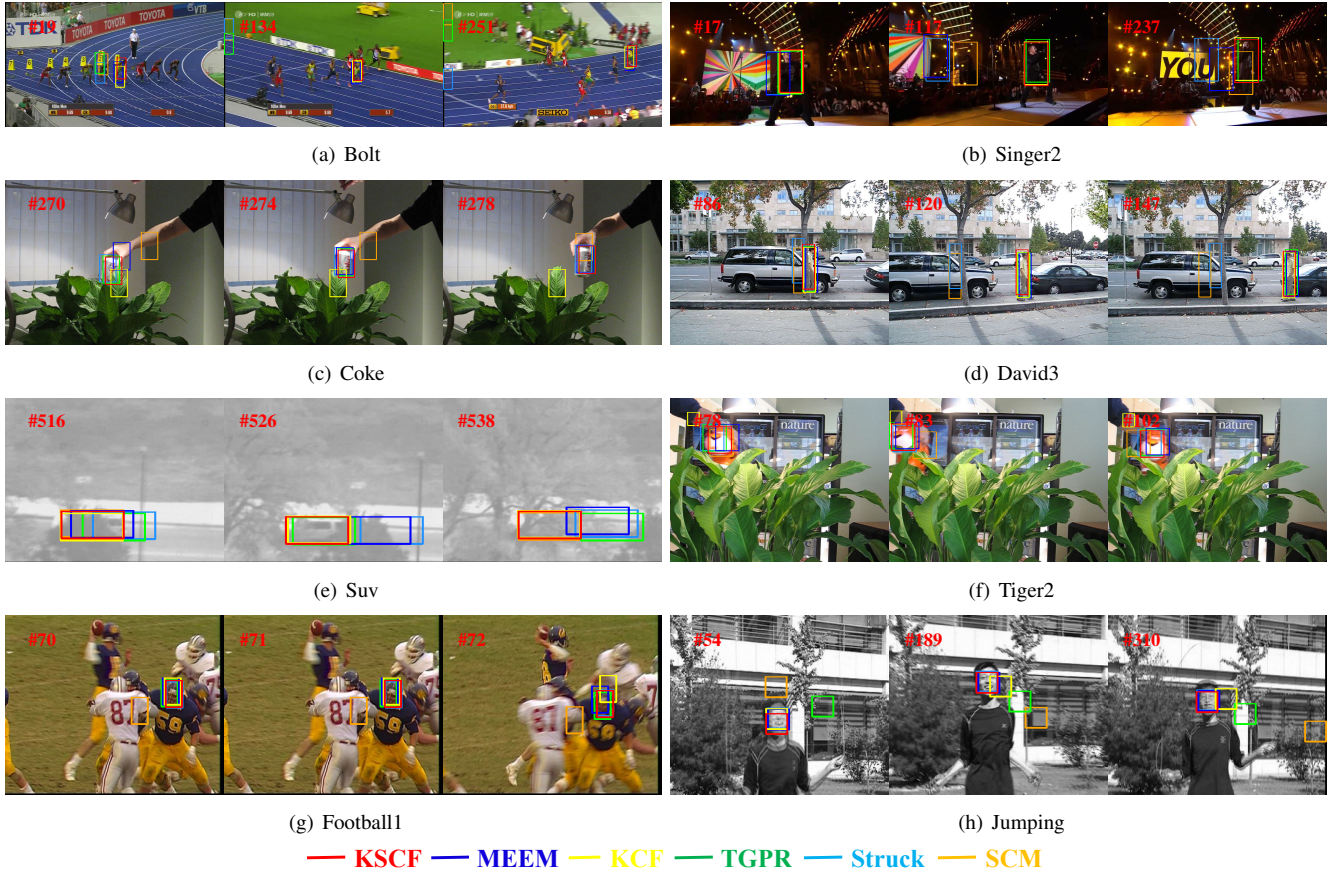
\begin{figure*}[htbp]
  \centering \small
  \subfigure[Bolt]{
    \begin{overpic}[width=0.5\textwidth]{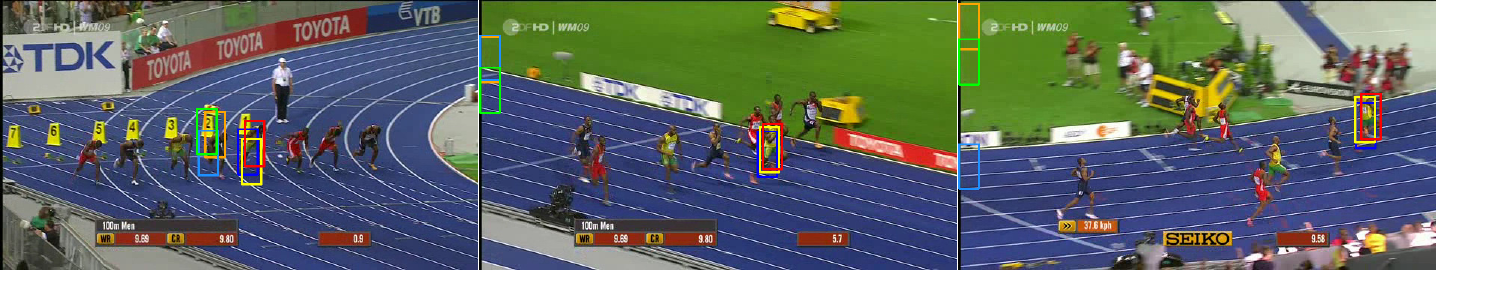}
    \label{fig:subfig:a} \scriptsize
    \put(0,15){\color{red} \ {\bf \#19}}
    \put(31,15){\color{red} \ {\bf \#134}}
    \put(63,15){\color{red} \ {\bf \#251}}
    \end{overpic}
  }
  \hspace{-0.26in}
  \subfigure[Singer2]{
    \begin{overpic}[width=0.5\textwidth]{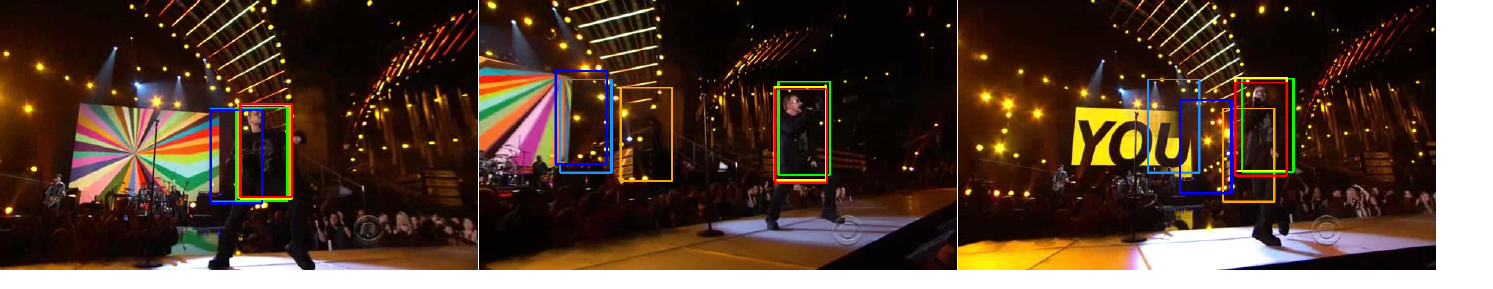}
    \label{fig:subfig:b} \scriptsize
    \put(0,15){\color{red} \ {\bf \#17}}
    \put(31,15){\color{red} \ {\bf \#117}}
    \put(63,15){\color{red} \ {\bf \#237}}
    \end{overpic}
  }
  \hspace{1in}
  \vspace{-0.05in}

  \subfigure[Coke]{
    \begin{overpic}[width=0.5\textwidth]{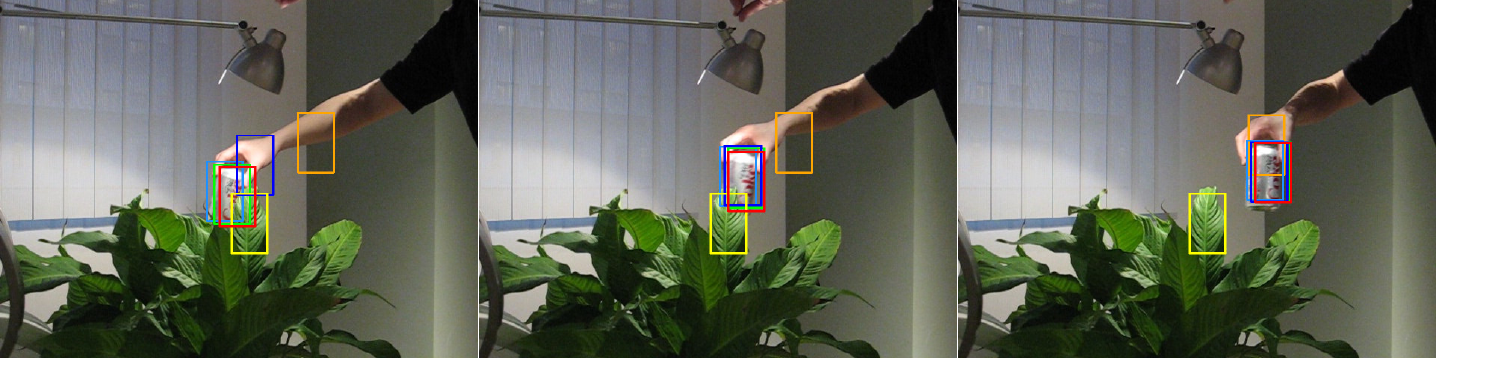}
    \label{fig:subfig:c} \scriptsize
    \put(0,21){\color{red} \ {\bf \#270}}
    \put(31,21){\color{red} \ {\bf \#274}}
    \put(63,21){\color{red} \ {\bf \#278}}
    \end{overpic}
    }
  \hspace{-0.26in}
  \subfigure[David3]{
    \begin{overpic}[width=0.5\textwidth]{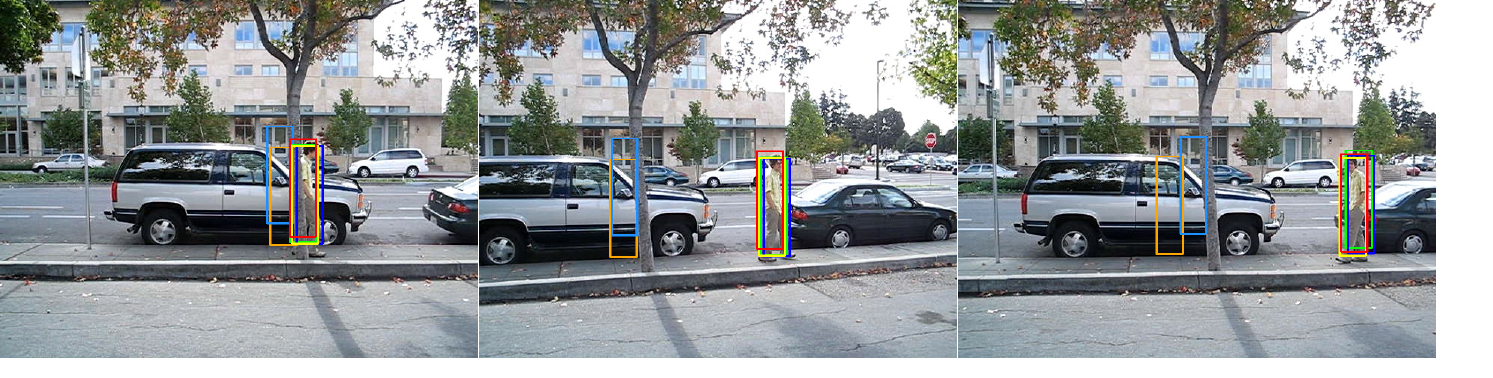}
    \label{fig:subfig:d} \scriptsize
    \put(0,21){\color{red} \ {\bf \#86}}
    \put(31,21){\color{red} \ {\bf \#120}}
    \put(63,21){\color{red} \ {\bf \#147}}
    \end{overpic}
    }
  \hspace{1in}
  \vspace{-0.05in}

  \subfigure[Suv]{
    \begin{overpic}[width=0.5\textwidth]{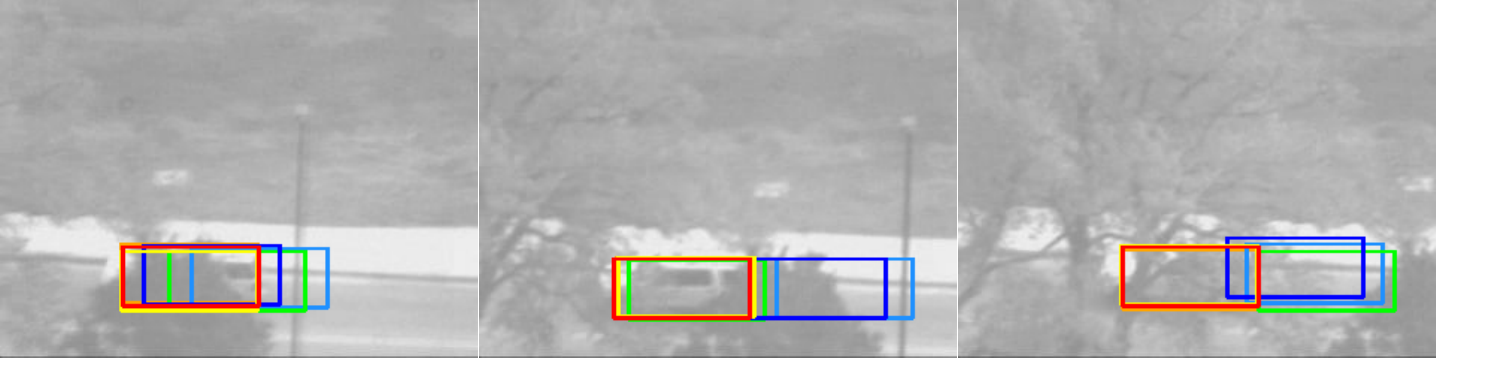}
    \label{fig:subfig:e} \scriptsize
    \put(0,21){\color{red} \ {\bf \#516}}
    \put(31,21){\color{red} \ {\bf \#526}}
    \put(63,21){\color{red} \ {\bf \#538}}
    \end{overpic}
    }
  \hspace{-0.26in}
  \subfigure[Tiger2]{
    \begin{overpic}[width=0.5\textwidth]{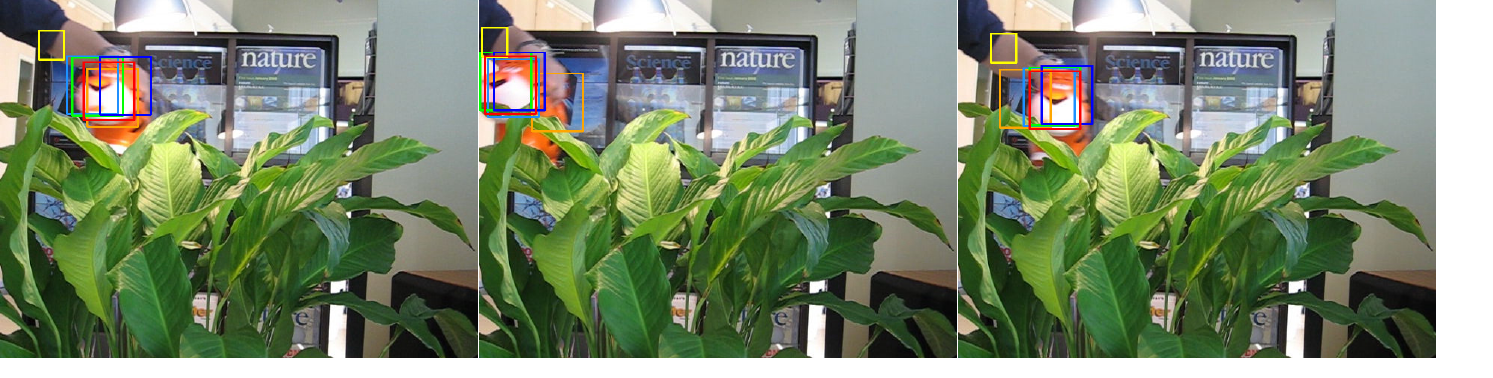}
    \label{fig:subfig:f} \scriptsize
    \put(0,21){\color{red} \ {\bf \#78}}
    \put(31,21){\color{red} \ {\bf \#83}}
    \put(63,21){\color{red} \ {\bf \#102}}
    \end{overpic}
    }
  \hspace{1in}
  \vspace{-0.05in}

  \subfigure[Football1]{
    \begin{overpic}[width=0.5\textwidth]{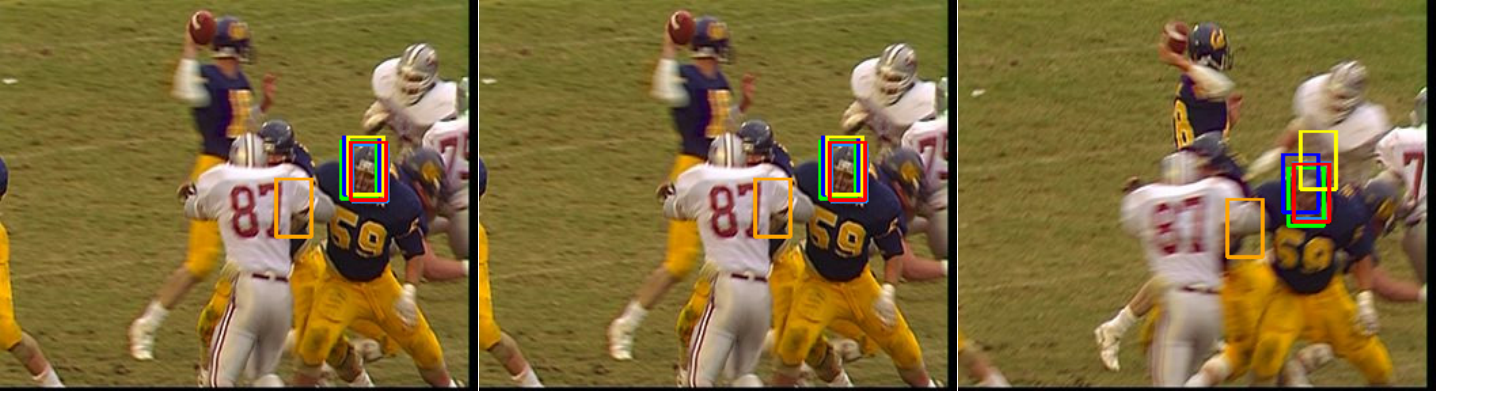}
    \label{fig:subfig:g} \scriptsize
    \put(0,23){\color{red} \ {\bf \#70}}
    \put(31,23){\color{red} \ {\bf \#71}}
    \put(63,23){\color{red} \ {\bf \#72}}
    \end{overpic}
    }
  \hspace{-0.26in}
  \subfigure[Jumping]{
    \begin{overpic}[width=0.5\textwidth]{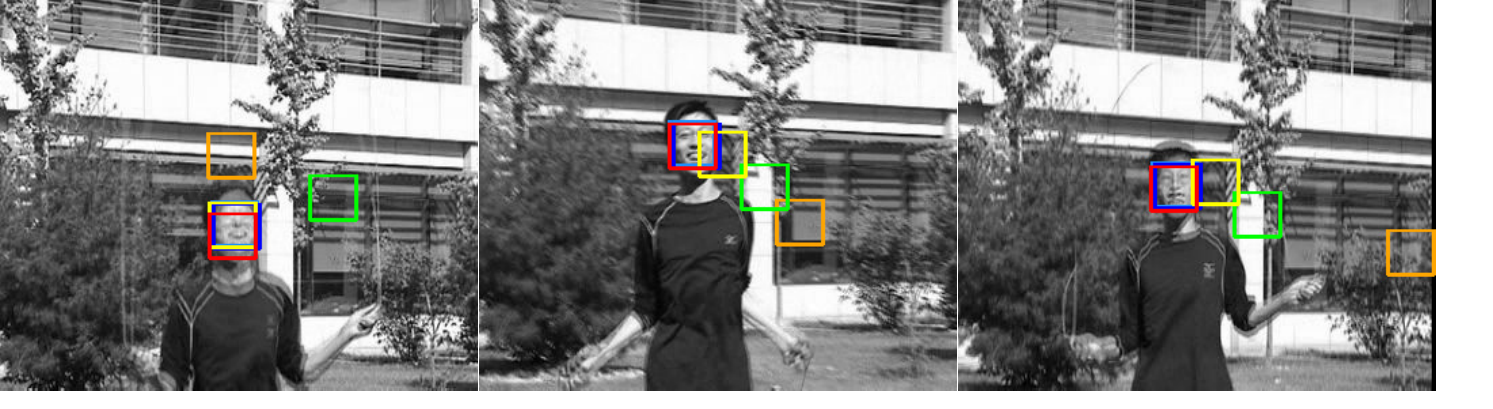}
    \label{fig:subfig:h} \scriptsize
    \put(0,23){\color{red} \ {\bf \#54}}
    \put(31,23){\color{red} \ {\bf \#189}}
    \put(63,23){\color{red} \ {\bf \#310}}
    \end{overpic}
    }
  \hspace{1in}
  \vspace{-0.05in}

    \centering
    \includegraphics[width=0.55\textwidth]{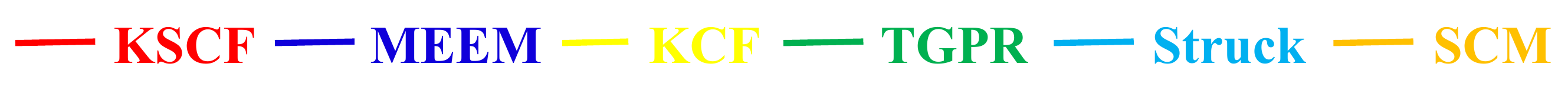}
  \vspace{-0.1in}
  \caption{Screenshots of tracking results on 8 challenging benchmark sequences. For the sake of clarity, we only show the results of five trackers, i.e., KSCF, KCF \cite{henriques2015high}, MEEM \cite{zhang2014meem}, TGPR \cite{gao2014transfer}, Struck \cite{hare2011struck} and SCM \cite{zhong2012robust}.}
  \label{fig:visual_evaluation} 
\end{figure*}
We evaluate the effect of kernel functions on KSCF using HOG and CN features,
%
including linear kernel  ${\bf {\it K}}_l({\bf x}_{i},{\bf x}_{j})={\bf x}^{\top}_{i}{\bf x}_{j}$,
polynomial kernel ${\bf {\it K}}_p({\bf x}_{i},{\bf x}_{j}) = ({\bf x}^{\top}_{i}{\bf x}_{j} + 1)^{\it d}$,
and Gaussian RBF kernel
${\bf {\it K}}_g({\bf x}_{i},{\bf x}_{j}) = \exp ( - \frac{1}{2 \sigma^2} \|{\bf x}_{i} - {\bf x}_{j} \|^2 )$.
For ${\bf {\it K}}_p({\bf x}_{i},{\bf x}_{j})$, the degree $d$ is set as $2$.
For ${\bf {\it K}}_g({\bf x}_{i},{\bf x}_{j})$, the kernel parameter $\sigma$ is set as $0.2$.
Table~\ref{table_KSCF_kernels} shows the results of KSCF with different kernels.
Clearly the KSCF method with a nonlinear kernel outperforms the one
with a linear kernel in terms of mean DP and mean AUC, and
the one with Gaussian RBF kernel achieves the best performance.

We implement the SKSCF method by extending KSCF with the Gaussian RBF kernel, and
compare four variants of the SCF-based trackers, i.e., SCF, MSCF, KSCF, and SKSCF.
Table \ref{table:scf-variants} shows the results of four SCF-based trackers,
where the SKSCF method performs best, followed by the KSCF approach.
On the other hand, the KSCF method is more efficient than the SKSCF approach.
%
In the following experiments,
we compare both KSCF and SKSCF methods with the other schemes based on
correlation filters, SVMs, and other state-of-the-art tracking approaches.
\subsection{Comparisons with CF-based trackers}
We use the tracking benchmark dataset \cite{wu2013online} to evaluate the
proposed SCF-based algorithm against existing CF-based methods
including MOSSE \cite{bolme2010visual}, CSK \cite{henriques2012exploiting}, KCF \cite{henriques2015high}, DCF \cite{henriques2015high}, STC \cite{zhang2014fast}, CN \cite{danelljan2014adaptive}, DSST \cite{danelljan2014accurate} and SAMF \cite{li2014scale}.
\begin{table*}[t] \small
\renewcommand{\arraystretch}{1}
\caption{Comparison of KSCF and SKSCF methods with the state-of-the-art trackers. The top three results are shown in red, blue and orange.}
\label{table1} \centering
\begin{tabular}{c|c|c|c|c|c|c|c|c|c|c|c}
\hline
{Algorithms} & \tabincell{c}{{SKSCF}\\{}}  & \tabincell{c}{{KSCF}\\{}}  & \tabincell{c}{MEEM \\ \cite{zhang2014meem}} & \tabincell{c}{KCF\\ \cite{henriques2015high}} & \tabincell{c}{TGPR\\ \cite{gao2014transfer}} & \tabincell{c}{SCM\\ \cite{zhong2012robust}} & \tabincell{c}{TLD\\ \cite{kalal2012tracking}} & \tabincell{c}{ASLA\\ \cite{jia2012visual}} & \tabincell{c}{L1APG\\ \cite{bao2012real}}  & \tabincell{c}{MIL\\ \cite{babenko2009visual}} & \tabincell{c}{CT\\ \cite{zhang2012real}} \\
\hline\hline
Mean DP (\%) & \textcolor[rgb]{1.00,0.00,0.00}{\bf 87.4} & \textcolor[rgb]{0.00,0.07,1.00}{\bf 85.0} & \textcolor[rgb]{1.00,0.50,0.00}{\bf 83.3} & 73.2 & 71.8 & 65.2 & 60.6 & 54.5 & 49.4 & 48.8 & 41.5\\
\hline
Mean AUC (\%) & \textcolor[rgb]{1.00,0.00,0.00}{\bf 62.3} & \textcolor[rgb]{0.00,0.07,1.00}{\bf 57.5} & \textcolor[rgb]{1.00,0.50,0.00}{\bf 57.2} & 50.7 & 51.1 & 50.1 & 43.4 & 44.2 & 38.6 & 36.9 & 30.8 \\
\hline
Mean FPS ({\it s}) & 8 & \textcolor[rgb]{1.00,0.50,0.00}{\bf 35} & 10 & \textcolor[rgb]{1.00,0.00,0.00}{\bf 172} & 0.5 & 1 & 22 & 8 & 3 & 28 & \textcolor[rgb]{0.00,0.07,1.00}{\bf 39} \\
\hline
\end{tabular}
\end{table*}
\begin{figure}[htbp]
\centering
\begin{tabular}{@{}c@{}c@{}}
 \includegraphics[scale=0.33]{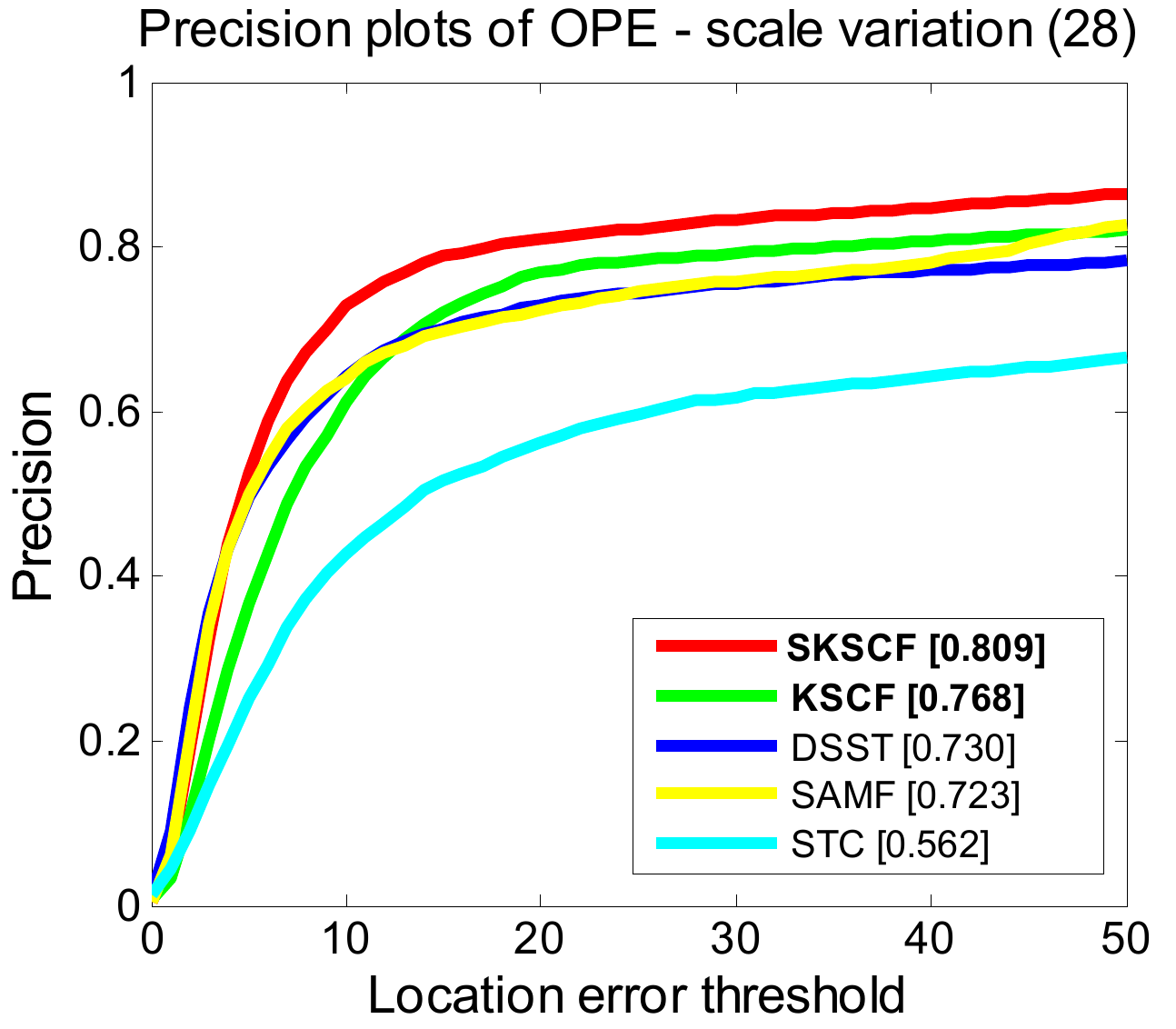} &
 \includegraphics[scale=0.33]{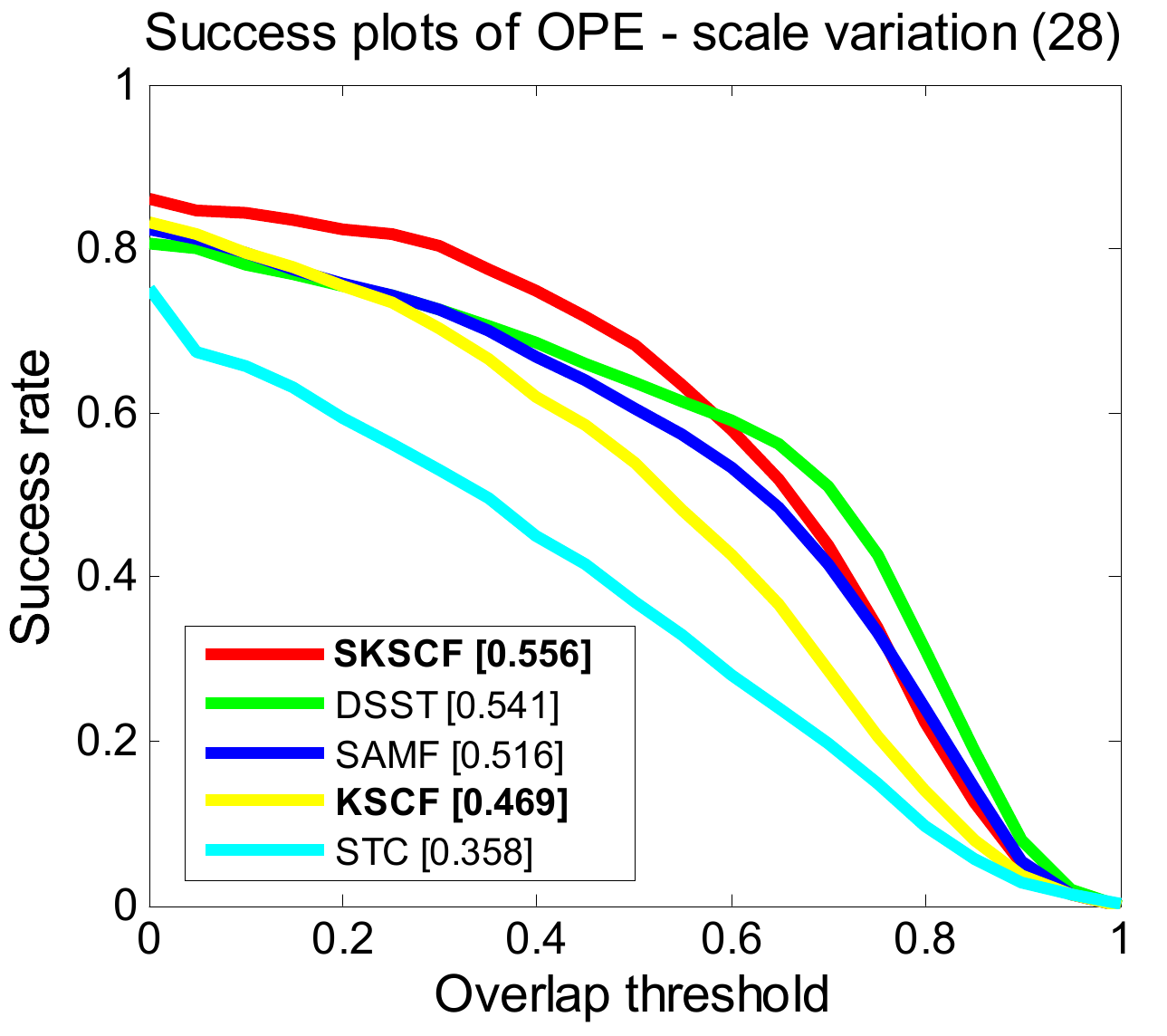}
\end{tabular}
\caption{OPE plots of the KSCF, SKSCF, DSST \cite{danelljan2014accurate} and SAMF \cite{li2014scale} methods on sequences with large scale variation.}
\label{fig:scale_attribute}
\end{figure}
\begin{figure}[htbp]
\centering
\begin{tabular}{@{}c@{}c@{}}
\includegraphics[scale=0.33]{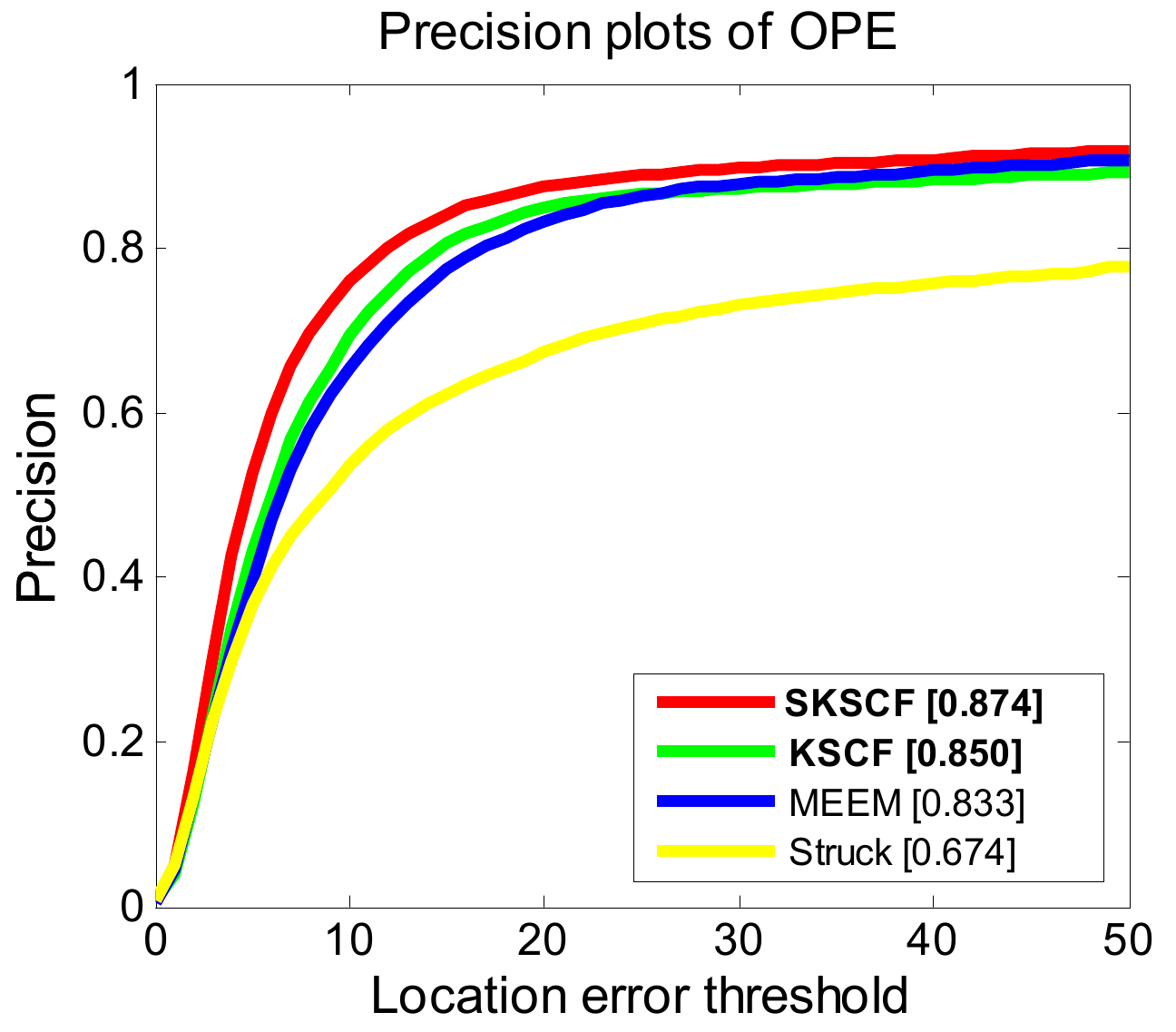} &
 \includegraphics[scale=0.33]{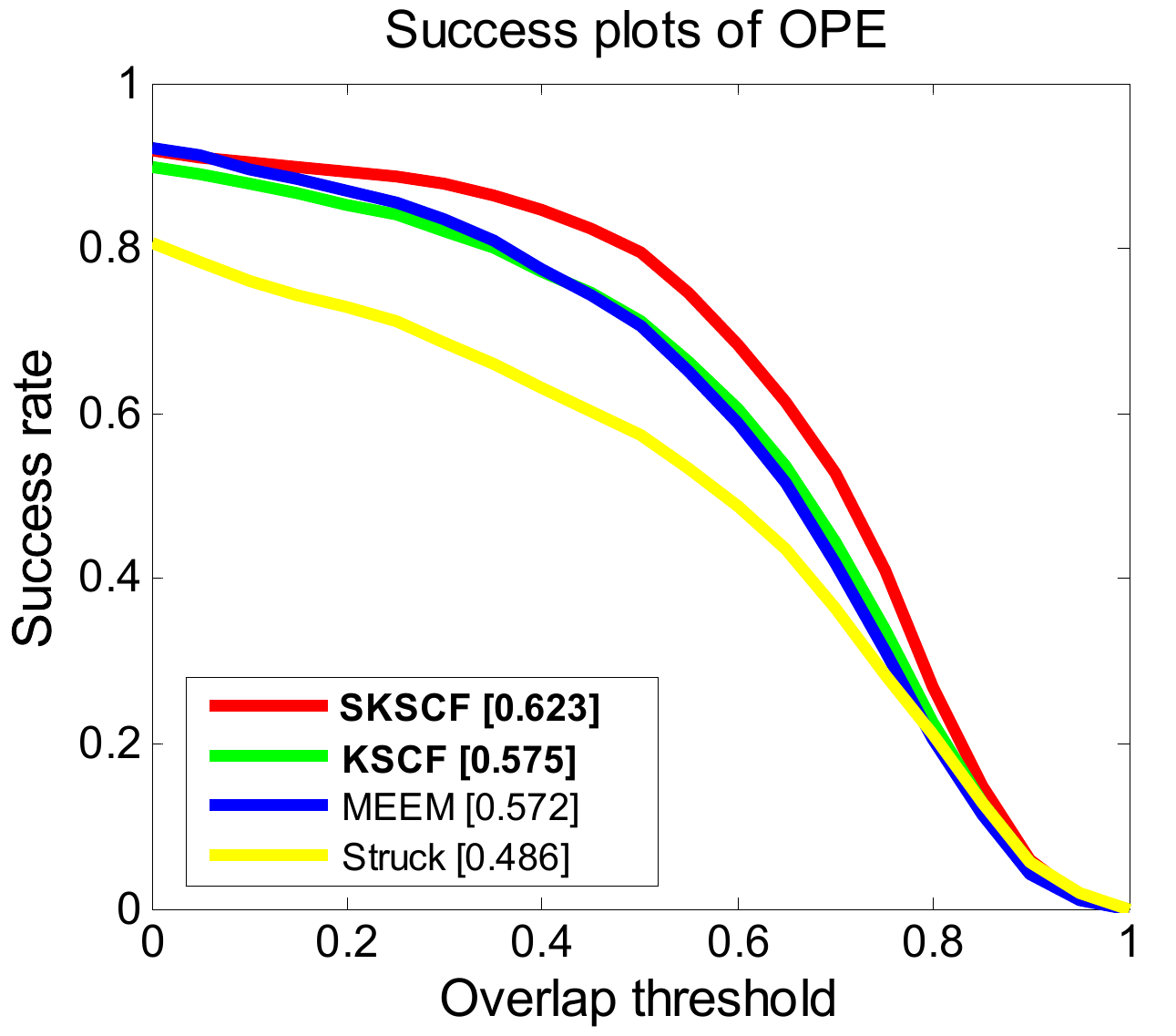}
 \end{tabular}
\caption{OPE plots of the KSCF, SKSCF and other SVM-based trackers, including MEEM \cite{zhang2014meem} and Struck \cite{hare2011struck}. }
\label{fig:svm_trackers}
\end{figure}
{\flushleft {\bf Classic correlation filters.}}
Fig.~\ref{fig:plots1} shows the OPE plots of these trackers.
The SCF, MOSSE \cite{bolme2010visual}, CSK \cite{henriques2012exploiting} and STC \cite{zhang2014fast} methods operate on raw pixels in the linear space.
We note that the MOSSE method adopts the ridge regression function while the SCF algorithm uses the
max-margin model.
Although the CSK and STC methods operate on raw pixels, the CSK method is a kernelized CF-based tracker and the STC approach is a scale-adaptive tracking method.
Overall, the SCF algorithm performs favorably against these CF-based methods based on regression and nonlinear functions.

{\flushleft {\bf Multi-channel correlation filters.}}
The MSCF, CN \cite{danelljan2014adaptive}, and DCF \cite{henriques2015high}
methods are based on correlation filters using multi-channel features.
The DCF method is based on HOG features and the CN approach is operated on color attributes,
while the MSCF scheme uses the combination of HOG and color representations.
Fig.~\ref{fig:plots1} shows that the MSCF method performs well among these three trackers based on correlation filters.
%

{\flushleft {\bf Kernelized correlation filters.}}
The KSCF method is compared with the corresponding kernelized KCF \cite{henriques2015high} and CSK \cite{henriques2012exploiting} trackers.
The CSK and KCF methods are based  on raw pixels and HOG features, respectively.
As shown in Table \ref{table:scf-variants} and Fig.~\ref{fig:visual_evaluation},
the KSCF method based on HOG and CN features performs favorably against the KCF and CSK
appraoches.
\begin{figure}[htbp]
\centering
\begin{tabular}{@{}c@{}c@{}}
\includegraphics[scale=0.33]{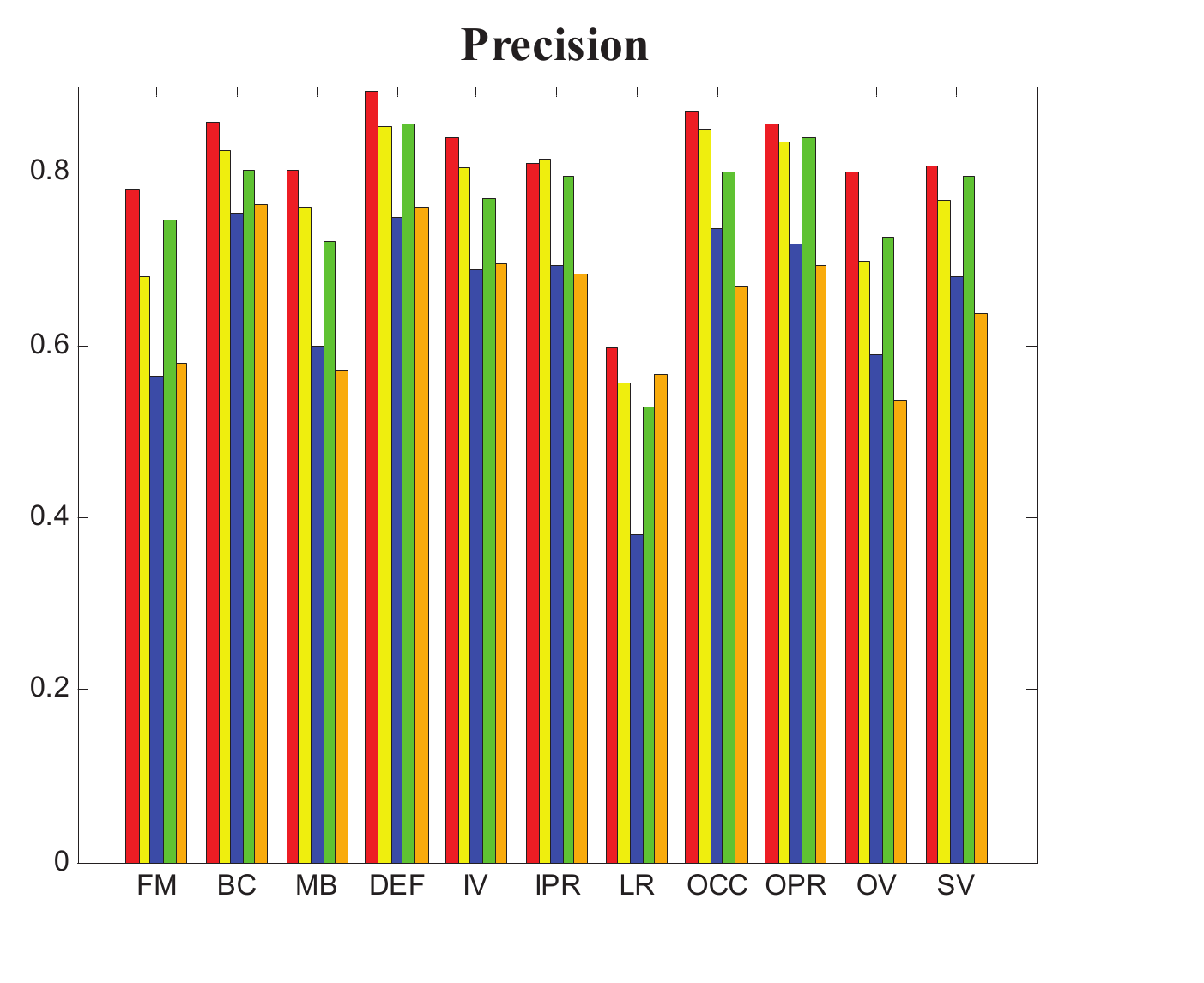} &
\hspace{-0.2in}
\includegraphics[scale=0.33]{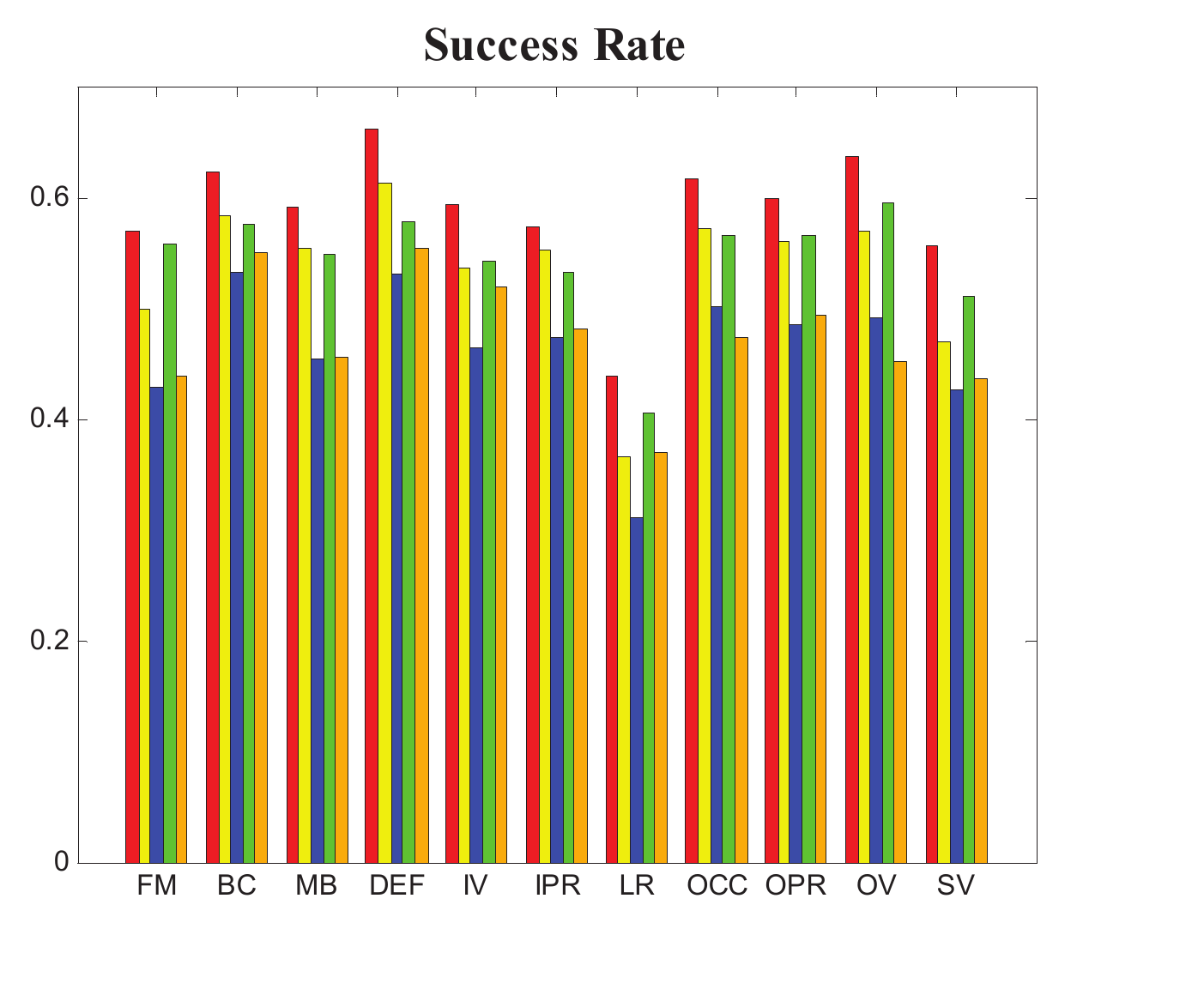}
\vspace{-0.1in}
\end{tabular}\\
\includegraphics[width=0.35\textwidth]{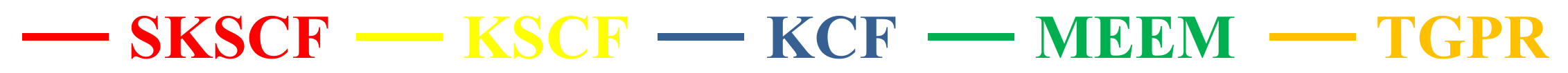}
\caption{Precision and success metrics of four top-performing  trackers for the 11 attributes.
}
\label{fig:bar_on_attributes}
\end{figure}
\begin{figure}[htbp]
\centering
\begin{tabular}{@{}c@{}c@{}}
\includegraphics[scale=0.33]{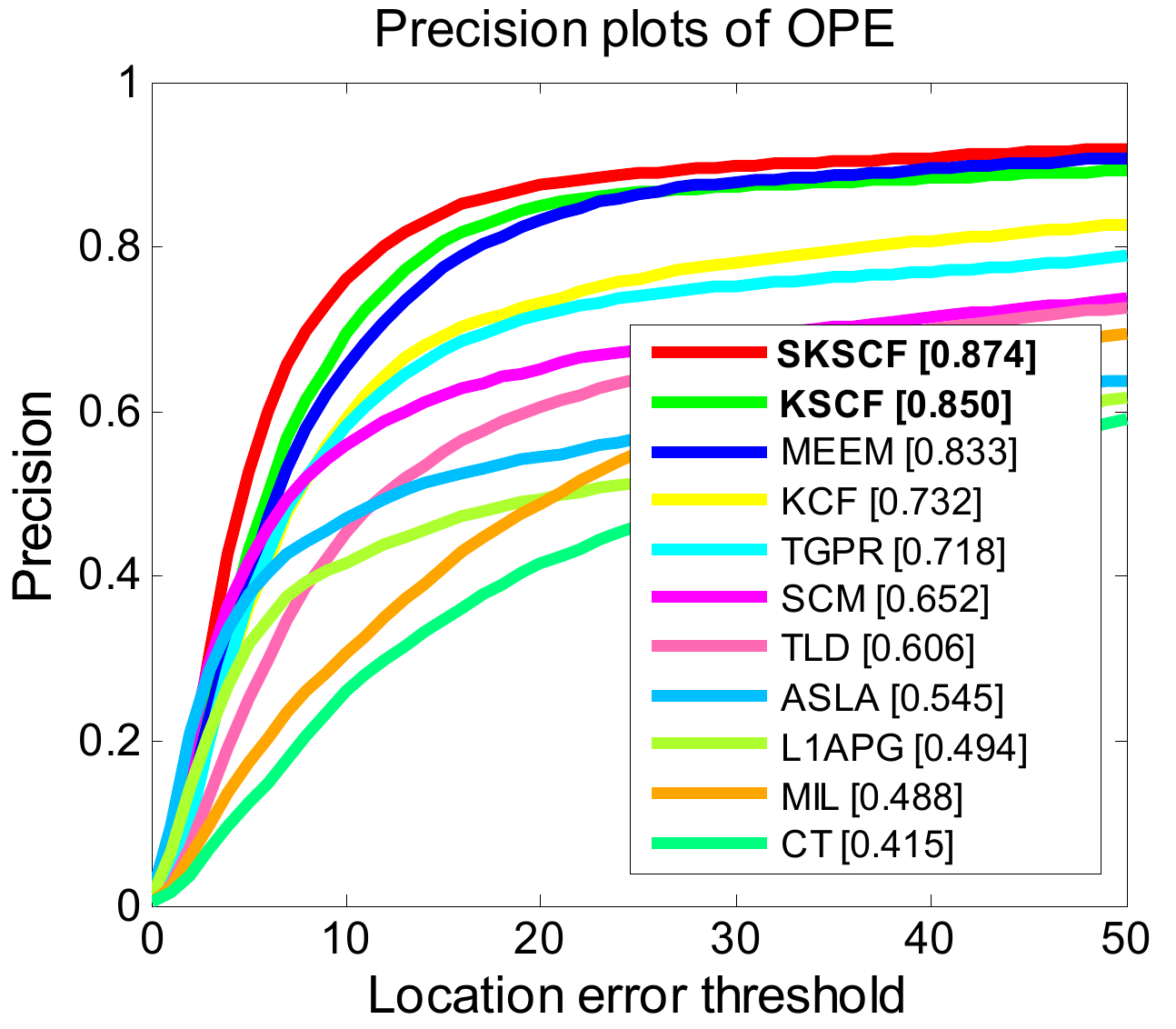} &
\includegraphics[scale=0.33]{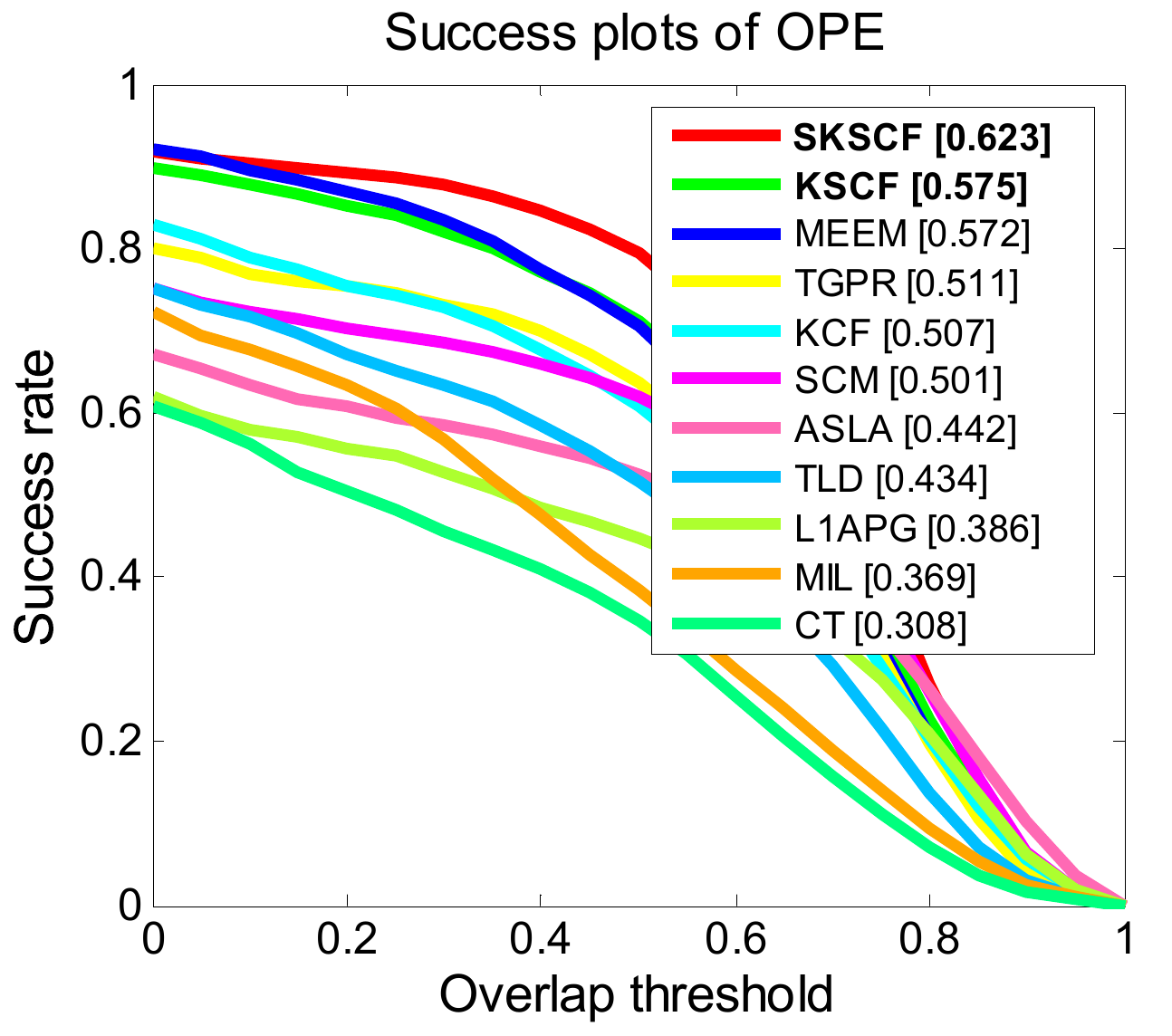}
\end{tabular}
\caption{OPE plots of the KSCF, SKSCF and other state-of-the art trackers, including MEEM \cite{zhang2014meem}, TGPR \cite{gao2014transfer}, KCF \cite{henriques2015high}, SCM \cite{zhong2012robust}, TLD \cite{kalal2012tracking}, ASLA \cite{jia2012visual}, L1APG \cite{bao2012real}, MIL \cite{babenko2009visual} and CT \cite{zhang2012real}.}
\label{fig:plots2}
\end{figure}
{\flushleft {\bf Scale-adaptive correlation filters.}}
The KSCF and SKSCF are evaluated against three scale-adaptive trackers:
STC \cite{zhang2014fast}, DSST \cite{danelljan2014accurate} and SAMF \cite{li2014scale}.
We note that
the DSST \cite{danelljan2014accurate} and SAMF \cite{li2014scale} methods have been shown to perform best and second best trackers in the recent tracking benchmark evaluation \cite{kristan2014visual}.
 Both KSCF and SKSCF trackers perform significantly better than the STC method.
In addition, the KSCF and SKSCF methods also significantly outperform the DSST  and SAMF approaches by a large margin.
Fig.~\ref{fig:scale_attribute} shows the OPE plots on all the sequences with the attribute of scale variation where  the KSCF method performs favorably against the DSST and SAMF trackers.
Overall, the KSCF algorithm performs favorably in terms of accuracy and speed.
\begin{table*}[t] \small
\renewcommand{\arraystretch}{1}
\caption{Precision metrics of the trackers for 11 attributes.
The top three results are shown in red, blue and orange.}
\label{table:precision_attributes} \centering
\begin{tabular}{c|c|c|c|c|c|c|c|c|c|c|c}
\hline
  Attributes      &   FM     &   BC    &   MB    &   DEF    &   IV    &   IPR    &   LR    &   OCC    &   OPR    &   OV    &   SV    \\
\hline\hline
{SKSCF} & \textcolor[rgb]{1.00,0.00,0.00}{\bf 0.779}   & \textcolor[rgb]{1.00,0.00,0.00}{\bf 0.859}   & \textcolor[rgb]{1.00,0.00,0.00}{\bf 0.802}   & \textcolor[rgb]{1.00,0.00,0.00}{\bf 0.893}    & \textcolor[rgb]{1.00,0.00,0.00}{\bf 0.841}  & \textcolor[rgb]{0.00,0.07,1.00}{\bf 0.810}   & \textcolor[rgb]{1.00,0.00,0.00}{\bf 0.596} & \textcolor[rgb]{1.00,0.00,0.00}{\bf 0.872}  & \textcolor[rgb]{1.00,0.00,0.00}{\bf 0.857}   & \textcolor[rgb]{1.00,0.00,0.00}{\bf 0.800}  & \textcolor[rgb]{1.00,0.00,0.00}{\bf 0.809}\\
\hline
{KSCF}                 & \textcolor[rgb]{1.00,0.50,0.00}{\bf 0.680}   & \textcolor[rgb]{0.00,0.07,1.00}{\bf 0.825}   & \textcolor[rgb]{0.00,0.07,1.00}{\bf 0.761}   & \textcolor[rgb]{1.00,0.50,0.00}{\bf 0.854}    & \textcolor[rgb]{0.00,0.07,1.00}{\bf 0.805}  & \textcolor[rgb]{1.00,0.00,0.00}{\bf 0.816}   & \textcolor[rgb]{1.00,0.50,0.00}{\bf 0.555} & \textcolor[rgb]{0.00,0.07,1.00}{\bf 0.852}  & \textcolor[rgb]{1.00,0.50,0.00}{\bf 0.836}   & \textcolor[rgb]{1.00,0.50,0.00}{\bf 0.697}  & \textcolor[rgb]{1.00,0.50,0.00}{\bf 0.768}\\
\hline
MEEM \cite{zhang2014meem}         & \textcolor[rgb]{0.00,0.07,1.00}{\bf 0.745}   & \textcolor[rgb]{0.00,0.07,1.00}{\bf 0.802}   & \textcolor[rgb]{0.00,0.07,1.00}{\bf 0.721}   & \textcolor[rgb]{0.00,0.07,1.00}{\bf 0.856}   & \textcolor[rgb]{0.00,0.07,1.00}{\bf 0.771}   & \textcolor[rgb]{0.00,0.07,1.00}{\bf 0.796}   & 0.529   & \textcolor[rgb]{0.00,0.07,1.00}{\bf 0.801}   & \textcolor[rgb]{0.00,0.07,1.00}{\bf 0.840}   & \textcolor[rgb]{0.00,0.07,1.00}{\bf 0.726}   & \textcolor[rgb]{0.00,0.07,1.00}{0.795}   \\
\hline
TGPR \cite{gao2014transfer}       & 0.579   & 0.763   & 0.570   & 0.760   & 0.695   & 0.683   & \textcolor[rgb]{0.00,0.07,1.00}{\bf 0.567}   & 0.668   & 0.693	& 0.535   & 0.637   \\
\hline
KCF \cite{henriques2015high}      & 0.564	& 0.752   & 0.599	& 0.747   & 0.687   & 0.692   & 0.379   & 0.735   & 0.718	& 0.589	  & 0.680   \\
\hline
SCM  \cite{zhong2012robust}       & 0.346	& 0.578	  & 0.358	& 0.589	  & 0.613	& 0.613	  & 0.305	& 0.646	  & 0.621	& 0.429	  & 0.672   \\
\hline
TLD   \cite{kalal2012tracking}    & 0.557	& 0.428   & 0.523   & 0.495	  & 0.540	& 0.588	  & 0.349	& 0.556	  & 0.593	& 0.576	  & 0.606   \\
\hline
ASLA  \cite{jia2012visual}        & 0.255   & 0.496   & 0.283   & 0.473   & 0.529   & 0.521   & 0.156   & 0.479   & 0.535   & 0.333   & 0.552   \\
\hline
L1APG \cite{bao2012real}          & 0.367  	& 0.425   & 0.379   & 0.398   & 0.341   & 0.524   & 0.460   & 0.475   & 0.490   & 0.329   & 0.472   \\
\hline
MIL   \cite{babenko2009visual}    & 0.415	& 0.456   & 0.381   & 0.493   & 0.359   & 0.465   & 0.171   & 0.448   & 0.484   & 0.393   & 0.471   \\
\hline
CT   \cite{zhang2012real}         & 0.330	& 0.339   & 0.314   & 0.463   & 0.365   & 0.361   & 0.152   & 0.429   & 0.405   & 0.336   & 0.448   \\
\hline
\end{tabular}
\end{table*}
\begin{table*}[t] \small
\renewcommand{\arraystretch}{1}
\caption{Success metrics of the trackers for 11 attributes.
The top three results are shown in red, blue and orange.}
\label{table:success_attributes} \centering
\begin{tabular}{c|c|c|c|c|c|c|c|c|c|c|c}
\hline
  Attributes      &   FM     &   BC    &   MB    &   DEF    &   IV    &   IPR    &   LR    &   OCC    &   OPR    &   OV    &   SV    \\
\hline\hline
{SKSCF}                  & \textcolor[rgb]{1.00,0.00,0.00}{\bf 0.729}   & \textcolor[rgb]{1.00,0.00,0.00}{\bf 0.795}   & \textcolor[rgb]{1.00,0.00,0.00}{\bf 0.757}   & \textcolor[rgb]{1.00,0.00,0.00}{\bf 0.863}    & \textcolor[rgb]{1.00,0.00,0.00}{\bf 0.743}  & \textcolor[rgb]{1.00,0.00,0.00}{\bf 0.720}   & \textcolor[rgb]{1.00,0.00,0.00}{\bf 0.542} & \textcolor[rgb]{1.00,0.00,0.00}{\bf 0.788}  & \textcolor[rgb]{1.00,0.00,0.00}{\bf 0.757}   & \textcolor[rgb]{1.00,0.00,0.00}{\bf 0.808}  & \textcolor[rgb]{1.00,0.00,0.00}{\bf 0.682}\\
\hline
{KSCF}                  & \textcolor[rgb]{1.00,0.50,0.00}{\bf 0.629}   & \textcolor[rgb]{1.00,0.50,0.00}{\bf 0.741}   & \textcolor[rgb]{1.00,0.50,0.00}{\bf 0.689}   & \textcolor[rgb]{0.00,0.07,1.00}{\bf 0.779}    & \textcolor[rgb]{1.00,0.50,0.00}{\bf 0.649}  & \textcolor[rgb]{0.00,0.07,1.00}{\bf 0.690}   & 0.389 & \textcolor[rgb]{0.00,0.07,1.00}{\bf 0.696}  & \textcolor[rgb]{0.00,0.07,1.00}{\bf 0.697}   & \textcolor[rgb]{1.00,0.50,0.00}{\bf 0.705}  & 0.540\\
\hline
MEEM \cite{zhang2014meem}         & \textcolor[rgb]{0.00,0.07,1.00}{\bf 0.706}   & \textcolor[rgb]{0.00,0.07,1.00}{\bf 0.747}   & \textcolor[rgb]{0.00,0.07,1.00}{\bf 0.692}   & \textcolor[rgb]{1.00,0.50,0.00}{\bf 0.711}     & \textcolor[rgb]{0.00,0.07,1.00}{\bf 0.653}   & \textcolor[rgb]{1.00,0.50,0.00}{\bf 0.648}   & \textcolor[rgb]{1.00,0.50,0.00}{\bf 0.470} & \textcolor[rgb]{1.00,0.50,0.00}{\bf 0.694}   & \textcolor[rgb]{1.00,0.50,0.00}{\bf 0.694}   & \textcolor[rgb]{0.00,0.07,1.00}{\bf 0.742}   & \textcolor[rgb]{1.00,0.50,0.00}{\bf 0.594}   \\
\hline
TGPR \cite{gao2014transfer}       & 0.542   & 0.713   & 0.570   & \textcolor[rgb]{1.00,0.50,0.00}{\bf 0.711}   & 0.632   & 0.601   & \textcolor[rgb]{0.00,0.07,1.00}{\bf 0.501}   & 0.592   & 0.603	& 0.546   & 0.505   \\
\hline
KCF \cite{henriques2015high}      & 0.516	& 0.669   & 0.539	& 0.668   & 0.534   & 0.575   & 0.358   & 0.593   & 0.587	& 0.589	  & 0.477   \\
\hline
SCM  \cite{zhong2012robust}       & 0.348	& 0.550	  & 0.358	& 0.566	  & 0.586	& 0.574	  & 0.308	& 0.602	  & 0.576	& 0.449	  & \textcolor[rgb]{0.00,0.07,1.00}{\bf 0.635}   \\
\hline
TLD   \cite{kalal2012tracking}    & 0.475	& 0.388   & 0.485   & 0.434	  & 0.461	& 0.477	  & 0.327	& 0.455	  & 0.489	& 0.516	  & 0.494   \\
\hline
ASLA  \cite{jia2012visual}        & 0.261   & 0.468   & 0.284   & 0.485   & 0.514   & 0.496   & 0.163   & 0.469   & 0.509   & 0.359   & 0.544   \\
\hline
L1APG \cite{bao2012real}          & 0.359  	& 0.404   & 0.363   & 0.398   & 0.298   & 0.445   & 0.458   & 0.437   & 0.423   & 0.341   & 0.407   \\
\hline
MIL   \cite{babenko2009visual}    & 0.353	& 0.414   & 0.261   & 0.440   & 0.300   & 0.339   & 0.157   & 0.378   & 0.369   & 0.416   & 0.335   \\
\hline
CT   \cite{zhang2012real}         & 0.327	& 0.323   & 0.262   & 0.420   & 0.308   & 0.290   & 0.143   & 0.360   & 0.325   & 0.405   & 0.342   \\
\hline
\end{tabular}
\end{table*}
\subsection{Comparisons with SVM-based trackers}
We evaluate the proposed KSCF and SKSCF with two state-of-the-art SVM-based methods,
 i.e., Struck \cite{hare2011struck} and MEEM \cite{zhang2014meem},
 based on the structured and ensemble learning.
Table~\ref{table:svm_based_trackers} and Fig. \ref{fig:svm_trackers} show that
both KSCF and SKSCF algorithms perform favorably against the MEEM and Struck methods in
all aspects.
%
%
As shown in Fig.~\ref{fig:visual_evaluation}, the KSCF algorithm can track target objects more precisely than other methods in the \textit{Singer2}, \textit{Coke}, \textit{Suv} and \textit{Tiger2} sequences.
%
The results show that dense sampling can be efficiently used with SVMs for effective visual tracking.
%
Fig.~\ref{fig:visual_evaluation} shows that the KSCF algorithm can track the objects more precisely in
all challenging sequences, while the other trackers tend to drift away from the target objects.
\subsection{Comparisons with state-of-the-art trackers}
We evaluate the KSCF algorithm with the other state-of-the-art trackers, including MEEM \cite{zhang2014meem}, KCF \cite{henriques2015high}, TGPR \cite{gao2014transfer}, SCM \cite{zhong2012robust}, TLD \cite{kalal2012tracking}, L1APG \cite{bao2012real}, MIL \cite{babenko2009visual}, ASLA \cite{jia2012visual} and CT \cite{zhang2012real}.
%
Fig.~\ref{fig:plots2} shows the OPE plots, and Table~\ref{table1} presents
the mean DP, AUC and FPS.
Overall, the proposed KSCF and SKSCF algorithms perform favorably against the state-of-the-art methods including the TLD, SCM, TGPR and MEEM schemes.

The sequences in the benchmark dataset \cite{wu2013online} are annotated with 11 challenging factors for visual tracking, including illumination variation (IV), scale variation (SV), occlusion (OCC), deformation (DEF), motion blur (MB), fast motion (FM), in-plane rotation (IPR), out-of-plane rotation (OPR), out-of-view (OV), background clutters (BC), and low resolution (LR).
Table \ref{table:precision_attributes} and Table \ref{table:success_attributes} show the performance of the
KSCF and state-of-the-art methods in terms of DP and AUC with respect to each factor.
Fig. \ref{fig:bar_on_attributes} shows the precision and success metrics of the
leading trackers (i.e., SKSCF, KSCF, MEEM, KCF and TGPR) with respect to the attributes.
We note that the MEEM method \cite{zhang2014meem}
adopts the multiple experts framework to deal with model drift
%
, and performs slightly better than KSCF for attributes FM, LR, OV and SV.
Overall, the KSCF algorithm are among the top 3 trackers for any attribute,
and the SKSCF algorithm performs best in both metrics for all but one attribute.
\section{Conclusions}
We propose an effective and efficient approach
to learn support correlation filters for real time visual tracking.
By reformulating the SVM model with circulant data matrix as training input,
we present a DFT based alternating optimization algorithm to learn support correlation filters
efficiently.
In addition, we develop the MSCF, KSCF, and SKSCF
tracking methods to exploit multidimensional features, nonlinear classifiers, and scale-adaptive schemes.
Experiments on a large benchmark dataset show that the proposed KSCF and SKSCF
algorithms perform favorably against the state-of-the-art tracking methods
in terms of accuracy and speed.

%

\section*{Acknowledgments}
This work is supported in part by NSFC grant (61271093), the program of ministry of education for new century excellent talents (NCET-12-0150), NSF CAREER Grant (No.
1149783) and NSF IIS Grant (No. 1152576).

\ifCLASSOPTIONcaptionsoff
  \newpage
\fi

\appendices
\section{Convergence analysis}
\label{Appendix_A}
%
%
In the following, we first analyze the optimality condition of the problem, and then prove the global convergence and convergence rate of the SCF algorithm.

\subsection{Optimality conditions}
In the spatial domain, the SCF model can be expressed as:
\begin{eqnarray} \label{eq:1}
\begin{aligned}
 &(\mathbf{w},b,\mathbf{e}) \rm{=} \arg \underset{\ \mathbf{w},b,\mathbf{e}}{\rm{\mathop{\min}}}\,
 \frac{1}{2}{{\left\| \mathbf{w} \right\|}^{2}}\rm{+}\frac{C}{2}{{\left\| \mathbf{y} \rm{\circ} ({{\mathbf{X}}^\top}\mathbf{w}\rm{+}b\mathbf{1})
 \rm{-}\mathbf{1}\rm{-}\mathbf{e} \right\|}^{2}},  \\
 & \mbox{s.t.} \  \mathbf{e}\ge 0 \nonumber
\end{aligned}
\end{eqnarray}
Defining the augmented vector $\mathbf{\tilde{x}}={{\left[ {{\mathbf{x}}^\top},1 \right]}^\top}$ with $\mathbf{x}\in{R^{n}}$, we compute
the augmented weight vector $\mathbf{\tilde{w}}={{\left[ {{\mathbf{w}}^\top},b \right]}^\top}$.
The above problem can then be reformulated as:
\begin{eqnarray} \label{eq:2}
\begin{aligned}
& (\mathbf{\tilde{w}},\mathbf{e})=\arg\ \underset{\mathbf{\tilde w,e}}{\mathop{\min}}\,\frac{1}{2}{{\mathbf{\tilde{w}}}^\top}\mathbf{\tilde{I}\tilde{w}}+\frac{C}{2}{{\left\| {{{\mathbf{\tilde{X}}}}^\top}\mathbf{\tilde{w}}-\mathbf{y}-\mathbf{y}\circ \mathbf{e} \right\|}^{2}}, \\
& \mbox{s.t.} \ \mathbf{e}\ge 0
\end{aligned}
\end{eqnarray}
where $\mathbf{\tilde{X}}={{\left[ {{\mathbf{X}}^\top},\mathbf{1} \right]}^\top}$ and $\mathbf{\tilde{I}} = \left[  \begin{aligned}
 & \mathbf{I}  \ & \ \mathbf{0} \\
 & \mathbf{0}^\top \ &  \ 0 \\
\end{aligned} \right]$. We introduce an indicator function
$\delta ({{e}_{i}})=\left\{ \begin{aligned}
  & -\infty \ ,\ \text{if}\ {{e}_{i}}<0 \\
 & {{e}_{i}}\ ,\ \ \text{if}\ {{e}_{i}}\ge 0 \\
\end{aligned} \right.\nonumber $
and the subdifferential \cite{rockafellar1970maximal} of $\delta ({{e}_{i}})$ is:
\begin{eqnarray} \label{eq:3}
\partial \delta ({{e}_{i}})=\left\{ \begin{aligned}
  & 1\ ,\ \ \ \ \ \ \ \ \ \ \ \ \ \ \ \ \ \ \text{if} \ {{e}_{i}}>0 \\
  & (-\infty ,0)\ ,\ \ \ \ \ \ \ \ \ \text{if} \ {{e}_{i}}=0 \\
  & \phi \mbox{(undefined)}\ ,\ \ \ \ \text{if}\ {{e}_{i}}<0 \\
\end{aligned} \right.
\end{eqnarray}
As the loss function (\ref{eq:2}) is convex,
$({{\mathbf{\tilde w}}^{*}},{{\mathbf{e}}^{*}})$ is a solution if and only if the subdifferential of the loss at $({{\mathbf{\tilde w}}^{*}},{{\mathbf{e}}^{*}})$ contains zero \cite{boyd2004convex}.
Thus the optimality conditions are:
\begin{eqnarray} \label{eq:4}
\left\{ \begin{aligned}
  & \mathbf{\tilde{w}\tilde{I}}+C\mathbf{\tilde{X}}({{{\mathbf{\tilde{X}}}}^{T}}\mathbf{\tilde{w}}-\mathbf{y}-\mathbf{y}\circ \mathbf{e})=0 \\
 & \left\{ \begin{aligned}
  & {{e}_{i}}+1-{{y}_{i}}\mathbf{\tilde{x}}_{i}^{T}\mathbf{\tilde{w}}=0\ ,\ \ \text{if}\ {{e}_{i}}>0 \\
 & {{y}_{i}}\mathbf{\tilde{x}}_{i}^{T}\mathbf{\tilde{w}}-1<0\ \ \ \ \ \ \ \ \text{if}\ {{e}_{i}}=0 \\
\end{aligned} \right. \\
\end{aligned} \right.
\end{eqnarray}
where $\mathbf{\tilde X}_{i}$ denotes the {\it i}-th training sample.
With $\lambda = \frac{1}{C}$, we have:
\begin{eqnarray}
\begin{aligned}
\nonumber  & \det \left( \lambda\mathbf{\tilde{I}}+\mathbf{\tilde{X}}{{{\mathbf{\tilde{X}}}}^\top} \right)=
\det \left( \left [ \begin{aligned}
  & \lambda\mathbf{{I}}+\mathbf{{X}}{{\mathbf{{X}}}^\top}\ \sum\limits_{i}{{{\mathbf{x}}_{i}}} \\
 & \sum\limits_{i}{\mathbf{x}_{i}^\top}\ \ \ \ \  \ \ {{n}^{2}} \\
\end{aligned} \right ]\right) \\
 & \ \ \ \ \ \ \ \ \ \ \ \ \ \ \ \ \ \ \ \ \ ={{n}^{2}}\det\left( \mathbf{X}{{\mathbf{X}}^\top}\!+\!\lambda \mathbf{I}\!-\!\frac{1}{{{n}^{2}}}\sum\limits_{i}{{{\mathbf{x}}_{i}}}\sum\limits_{i}{\mathbf{x}_{i}^\top} \right) \\
 & \ \ \ \ \ \ \ \ \ \ \ \ \ \ \ \ \ \ \ \ \ ={{n}^{2}}\det\left( {{\mathbf{X}}_c}\mathbf{X}_c^\top+\lambda \mathbf{I} \right),
 \\
\end{aligned}
\end{eqnarray}
where $\mathbf{X}_c=\left [ \mathbf{{x}_{1}}-\mathbf{\bar x}, \ldots, \mathbf{{x}_{n}}-\mathbf{\bar x} \right ]$
with $\mathbf{\bar x}=\frac{1}{n}\sum\limits_{i}{\mathbf{x}_{i}}$.
Thus the matrix ${{(\lambda\mathbf{\tilde I}+\mathbf{\tilde X}{{\mathbf{\tilde X}}^\top})} }$ is invertible.
For simplicity, let ${\mathbf M}=\mathbf{\tilde I}+C{\mathbf{\tilde X}}{{\mathbf{\tilde X}}}^\top$,
from (\ref{eq:4}) and above equation, we have
\begin{eqnarray}
\label{eq:5}
\mathbf{\tilde{w}}=C{\mathbf{M}^{-1}}\mathbf{\tilde{X}}\left( \mathbf{y}+\mathbf{y}\circ \mathbf{e} \right),
\end{eqnarray}
%
\begin{eqnarray}
\label{eq:6}
{{(C\mathbf{y}\circ{\mathbf{M}^{-1}}\mathbf{\tilde{X}}(\mathbf{y}+\mathbf{y}\circ \mathbf{e})-1)}_{i}} \left\{ \begin{aligned}
  & \!=\!{{e}_{i}},\ \ \text{if}\ \!{{e}_{i}}\!>\!0 \\
 & \!<\!0,\ \ \ \text{if}\ \!{{e}_{i}}\!=\!0\  \\
\end{aligned} \right.
\end{eqnarray}
Based on the optimality conditions in (\ref{eq:4}), we define
\begin{eqnarray} \label{eq:7}
\left\{ \begin{aligned}
\nonumber  & {{r}_{1}}=\mathbf{\tilde w}+C\mathbf{\tilde X}({{\mathbf{\tilde X}}^\top}\mathbf{\tilde w}-\mathbf{y}-\mathbf{y}\circ \mathbf{e}), \\
 & {{r}_{2}}(i)={{e}_{i}}+1-{{y}_{i}}\mathbf{\tilde x}_{i}^\top\mathbf{\tilde w} \ \ \forall {{e}_{i}}>0, \\
 & {{r}_{3}}(i)={{y}_{i}}\mathbf{\tilde x}_{i}^\top\mathbf{\tilde w}-1 \  \ \forall {{e}_{i}}\le 0, \\
\end{aligned} \right.
\end{eqnarray}
and use the stopping criterion:
\begin{eqnarray} \label{eq:8}
\begin{aligned}
\nonumber \max \left\{ {{\left\| {{r}_{1}} \right\|}_{\infty }}, \underset{{{e}_{i}}>0}{\mathop{\max }}\,\left\| {{r}_{2}}(i) \right\|, \underset{{{e}_{i}}=0}{\mathop{\max }}\,\left\| {{r}_{3}}(i) \right\| \right\}\le \epsilon,
\end{aligned}
\end{eqnarray}
where $\epsilon >0$ is a predefined threshold.

\subsection{Global convergence}
To compute $\mathbf{e}$, we reformulate
the sub-problem for each entry:
\begin{eqnarray} \label{eq:9}
\begin{aligned}
\nonumber \hat{z}=\arg \underset{z}{\mathop{\min }}\,\ \frac{1}{2}{{\left\| z-{{z}_{0}} \right\|}^{2}}+\delta (z),
\end{aligned}
\end{eqnarray}
where $\delta (z)=\left\{ \begin{aligned}
  & -\infty ,\  \text{if}\ z<0 \\
 & z,\ \ \ \ \  \ \text{if}\ z\ge 0 \\
\end{aligned} \right.$.
Its solution is given by:
\begin{eqnarray}
\label{eq:10}
\nonumber \hat{z}=g({{z}_{0}})=\left\{ \begin{aligned}
  & {{z}_{0}},\ \ \text{if}\ {{z}_{0}}\ge 0 \\
 & 0,\ \ \ \text{if}\ {{z}_{0}}<0 \\
\end{aligned} \right.
\end{eqnarray}

\begin{Proposition}
\label{Proposition_1}
For any $a, b\in R$, we have:
\begin{eqnarray} \label{eq:11}
\nonumber {{\left\| g(a)-g(b) \right\|}^{2}}\le {{\left\| a-b \right\|}^{2}},
\end{eqnarray}
where the equality holds only if $g(a)-g(b)=a-b$.
\begin{proof}
\
\begin{enumerate}
    \item if $a, b\ge 0$, ${{\left\| g(a)-g(b) \right\|}^{2}}={{\left\| a-b \right\|}^{2}}$, and we also have $g(a)-g(b)=a-b$.

    \item if $a, b<0$, ${{\left\| g(a)-g(b) \right\|}^{2}}=0\le {{\left\| a-b \right\|}^{2}}$, where the equality holds only if $a=b$.

    \item if $ab<0$, e.g., $b<0$, it is easy to see that, ${{a}^{2}}<{{\left( \left| a \right|+\left| b \right| \right)}^{2}}$.
\end{enumerate}
\end{proof}
\end{Proposition}
For simplicity, let $\mathbf{U}=\mathbf{\tilde X}\mbox{Diag}(\mathbf{y})$.
We have $\mathbf{U}{{\mathbf{U}}^\top}=\mathbf{\tilde X}{{\mathbf{\tilde X}}^\top}$ and then we get two
symmetric positive definite matrices as follows:
\begin{eqnarray}
\nonumber {\mathbf M}=\mathbf{\tilde I}+C{\mathbf{\tilde X}}{{\mathbf{\tilde X}}}^\top=\mathbf{\tilde I}+C{\mathbf U}{{\mathbf U}}^\top,
\end{eqnarray}
\begin{eqnarray}
\nonumber \mathbf{T}=C{{\mathbf{U}}^\top}{{(\mathbf{\tilde I}+C\mathbf{U}{{\mathbf{U}}^\top})}^{-1}}\mathbf{U}=C{{\mathbf{U}}^\top}{{\mathbf{M}}^{-1}}\mathbf{U},
\end{eqnarray}
where $\rho (\mathbf{T})<1$ and $\rho (\mathbf{T})$ is the spectral radius of matrix $\mathbf T$ \cite{horn2012matrix}.

With the definitions of $\mathbf{M}$ and $\mathbf{T}$, the updating rules $\mathbf{\tilde w}$ and $\mathbf{e}$ can be written as:
\begin{eqnarray}\label{eq:12}
\nonumber {{\mathbf{e}}^{k+1}}=g({{\mathbf{U}}^\top}{{\mathbf{\tilde w}}^{k}}-1)=g(\mathbf{T}(\mathbf{1}+{{\mathbf{e}}^{k}})-1)=g\circ h({{\mathbf{e}}^{k}}),
\end{eqnarray}
\begin{eqnarray}\label{eq:13}
\nonumber {{\mathbf{\tilde w}}^{k+1}}=C{{\mathbf{M}}^{-1}}\mathbf{U}(\mathbf{1}+{{\mathbf{e}}^{k+1}}),
\end{eqnarray}
Let $h({{\mathbf{e}}^{k}})=\mathbf{T}(\mathbf{1}+{{\mathbf{e}}^{k}})-1$, we have the following proposition.

\begin{Proposition}
\label{Proposition_2}
For any $\mathbf{e}\ne \mathbf{\hat{e}}$, the following inequality holds:
\begin{eqnarray} \label{eq:14}
\nonumber \left\| h(\mathbf{e})-h(\mathbf{\hat{e}}) \right\|\le \left\| \mathbf{e}-\mathbf{\hat{e}} \right\|,
\end{eqnarray}
and the equality holds if and only if $h(\mathbf{e})-h(\mathbf{\hat{e}})=\mathbf{e}-\mathbf{\hat{e}}$.
\begin{proof}
Note that $\rho (\mathbf{T})<1$. From the definition of $h(\mathbf{e})$, we have:
\begin{eqnarray}
\nonumber \left\| h(\mathbf{e})-h(\mathbf{\hat{e}}) \right\|=\left\| \mathbf{T}(\mathbf{e}-\mathbf{\hat{e}}) \right\|\le \rho (\mathbf{T})\left\| \mathbf{e}-\mathbf{\hat{e}} \right\|<\left\| \mathbf{e}-\mathbf{\hat{e}} \right\|,
\end{eqnarray}
Denote the eigen-decomposition of $\mathbf{T}$ by $\mathbf{T=}{{\mathbf{Q}}^\top}\mathbf{\Lambda Q}$, where $\mathbf{Q}$ is a full rank orthogonal matrix, and $\mathbf{\Lambda }$ is a diagonal matrix with $0\le {{\lambda }_{i}}\le 1$.

The equality $\left\| h(\mathbf{e})\!-\!h(\mathbf{\hat{e}}) \right\|\!\!=\!\!\left\| \mathbf{e}\!-\!\mathbf{\hat{e}} \right\|$ can be written as $\left\| {{\mathbf{Q}}^\top}\mathbf{\Lambda Q}(\mathbf{e\!-\!\hat{e}}) \right\|\!=\!\left\| \mathbf{e}\!-\!\mathbf{\hat{e}} \right\|$. Since $\mathbf{Q}$ is full-rank orthogonal, there is $\left\| \mathbf{e}-\mathbf{\hat{e}} \right\|=\left\| \mathbf{Q}(\mathbf{e}-\mathbf{\hat{e}}) \right\|$. Thus, we have $\left\| \mathbf{\Lambda Q}(\mathbf{e}-\mathbf{\hat{e}}) \right\|=\left\| \mathbf{Q}(\mathbf{e}-\mathbf{\hat{e}}) \right\|$. Since $\mathbf{\Lambda }$ is diagonal with $0\le {{\lambda }_{i}}\le 1$, it holds that $\mathbf{\Lambda Q}(\mathbf{e}-\mathbf{\hat{e}})=\mathbf{Q}(\mathbf{e}-\mathbf{\hat{e}})$. Multiplying both sides by ${{\mathbf{Q}}^\top}$, we have $\mathbf{T}(\mathbf{e}-\mathbf{\hat{e}})=h(\mathbf{e})-h(\mathbf{\hat{e}})=\mathbf{e}-\mathbf{\hat{e}}$.
\end{proof}
\end{Proposition}

\begin{Definition}(Fixed point) \cite{goebel1972fixed}
\label{Definition_Fixed_point}
Given a linear operator, a point ${{x}^{*}}$ is a fixed point if ${{x}^{*}}=f({{x}^{*}})$.
We next provide the following property for fixed points of the operator $g\circ h$.
\end{Definition}

\begin{Lemma}
\label{Lemma_3}
Given any fixed point $\mathbf{\hat{e}}$ of $g\circ h$, for any $\mathbf{e}$, we have:
\begin{eqnarray}
\nonumber \left\| g\circ h(\mathbf{e})-g\circ h(\mathbf{\hat{e}}) \right\|<
\left\| \mathbf{e}-\mathbf{\hat{e}} \right\|,
\end{eqnarray}
unless $\mathbf{e}$ is a fixed point of $g\circ h$.
\begin{proof}
From Propositions~\ref{Proposition_1} and ~\ref{Proposition_2}, it holds:
\begin{eqnarray}
\nonumber \left\| g\circ h(\mathbf{e})-g\circ h(\mathbf{\hat{e}}) \right\| <
\left\| h(\mathbf{e})-h(\mathbf{\hat{e}}) \right\|<\left\| \mathbf{e}-\mathbf{\hat{e}} \right\|,
\end{eqnarray}
unless $g\circ h(\mathbf{e})-g\circ h(\mathbf{\hat{e}})=h(\mathbf{e})-h(\mathbf{\hat{e}})=\mathbf{e}-\mathbf{\hat{e}}$. Thus if $g\circ h(\mathbf{\hat{e}})=\mathbf{\hat{e}}$, we have $g\circ h(\mathbf{e})=\mathbf{e}$.
\end{proof}
\end{Lemma}

\begin{Theorem}(Global convergence)
\label{Theorem_4}
The sequence $\left\{ ({{\mathbf{\tilde w}}^{k}},{{\mathbf{e}}^{k}}) \right\}$ generated by our algorithm from any starting point $({{\mathbf{\tilde w}}^{0}},{{\mathbf{e}}^{0}})$ converges to a solution $({{\mathbf{\tilde w}}^{*}},{{\mathbf{e}}^{*}})$ of the optimization problem.
\begin{proof}
First we prove that ${{\mathbf{e}}^{k}}$ converges to a fixed point.
Note that $g\circ h$ is non-expansive, thus the sequence $\{{{\mathbf{e}}^{k}}\}$ lies in a compact region and ${{\mathbf{e}}^{k}}$  converges to one limit point ${{\mathbf{e}}^{*}}$ at least.
We assume ${{\mathbf{e}}^{*}}=\underset{j\to \infty }{\mathop{\lim }}\,{{\mathbf{e}}^{{{k}_{j}}}}$ and let $\mathbf{\hat{e}}$ be any fixed point of $g\circ h$ with $\mathbf{\hat{e}}=g\circ h(\mathbf{\hat{e}})$.
Then the following formula is established:
\begin{eqnarray} \label{eq:15}
\nonumber \left\| {{\mathbf{e}}^{k}}-\mathbf{\hat{e}} \right\|=\left\| g\circ h({{\mathbf{e}}^{k-1}})-g\circ h(\mathbf{\hat{e}}) \right\|\le \left\| {{\mathbf{e}}^{k-1}}-\mathbf{\hat{e}} \right\|,
\end{eqnarray}
Based on above, we get the limit as below:
\begin{eqnarray} \label{eq:16}
\nonumber \underset{k\to \infty }{\mathop{\lim }}\,\left\| {{\mathbf{e}}^{k}}-\mathbf{\hat{e}} \right\|=\underset{j\to \infty }{\mathop{\lim }}\,\left\| {{\mathbf{e}}^{{{k}_{j}}}}-
\mathbf{\hat{e}} \right\|=\left\| {{\mathbf{e}}^{*}}-\mathbf{\hat{e}} \right\|,
\end{eqnarray}
which shows that more than one of all limit points of $\{{{\mathbf{e}}^{k}}\}$ have an equal distance to $\mathbf{\hat{e}}$. Because of the continuity of $g\circ h$, we have:
%
\begin{align}
\label{eq:17}
\nonumber g\circ h({{\mathbf{e}}^{*}}) =\underset{j\to \infty }{\mathop{\lim }}\,g\circ h({{\mathbf{e}}^{{{k}_{j}}}})=\underset{j\to \infty }{\mathop{\lim }}\,{{\mathbf{e}}^{{{k}_{j}}+1}}.
\end{align}
Thus, $g\circ h({{\mathbf{e}}^{*}})$ is also a limit point of sequence $\{{{\mathbf{e}}^{k}}\}$ and
it must have an equal distance to $\mathbf{\hat{e}}$:
\begin{eqnarray} \label{eq:18}
\nonumber \left\| {{\mathbf{e}}^{*}}-\mathbf{\hat{e}} \right\|=\left\| g\circ h({{\mathbf{e}}^{*}})-\mathbf{\hat{e}} \right\|=\left\| g\circ h({{\mathbf{e}}^{*}})-g\circ h(\mathbf{\hat{e}}) \right\|
\end{eqnarray}
According to Lemma ~\ref{Lemma_3}, we know $g\circ h({{\mathbf{e}}^{*}})={{\mathbf{e}}^{*}}$. Since $\mathbf{\hat{e}}$ is any fixed point of $g\circ h$, with the continuity of $g\circ h({{\mathbf{e}}^{*}})$, the convergence: $\underset{k\to \infty }{\mathop{\lim }}\,{{\mathbf{e}}^{k}}={{\mathbf{e}}^{*}}$ is obtained.
We next show that ${{\mathbf{e}}^{*}}$ satisfies the optimization condition in
(\ref{eq:6}).
With the definition of $\mathbf{T}$, $g$ and $h$, we have:
\begin{align}
\nonumber
& g\circ h(\mathbf{e}) =g(\mathbf{T}(\mathbf{1}+\mathbf{e})-1) \\
 & \rm{=}\ g(C{{\mathbf{U}}^\top}{{(\mathbf{\tilde I}\rm{+}C\mathbf{U}{{\mathbf{U}}^\top})}^{\!-\!1}}\mathbf{U}(\mathbf{1}\rm{+}\mathbf{e})\rm{-}1) \left\{ \begin{aligned}
  & \! \rm{=} {{e}_{i}}, \text{if}\ {{e}_{i}}\rm{>}0 \\
 & \! \rm{<} 0,\ \text{if}\ {{e}_{i}}\rm{=}0\  \\
\end{aligned} \right.,
\end{align}
which can be written as $\mathbf{e}=g\circ h(\mathbf{e})$.
Considering $g\circ h({{\mathbf{e}}^{*}})={{\mathbf{e}}^{*}}$, the solution ${{\mathbf{e}}^{*}}$ satisfies
the optimization conditions and the proposed algorithm converges to the global optimum.
\end{proof}
\end{Theorem}

\subsection{q-linear convergence rate}
\begin{Theorem}(Convergence rate)
\label{Theorem_5}
The sequence $\left\{ ({{\mathbf{\tilde w}}^{k}},{{\mathbf{e}}^{k}}) \right\}$ generated by our algorithm satisfies the following 3 conditions:
\begin{enumerate}
    \item $\left\| {{\mathbf{e}}^{k+1}}-{{\mathbf{e}}^{*}} \right\|\le \sqrt{\rho ({{\mathbf{T}}^{2}})}\left\| {{\mathbf{e}}^{k}}-{{\mathbf{e}}^{*}} \right\|$,

    \item $\left\| {{\mathbf{U}}^\top}({{\mathbf{\tilde w}}^{k+1}}-{{\mathbf{\tilde w}}^{*}}) \right\|\le \sqrt{\rho ({{\mathbf{T}}^{2}})}\left\| {{\mathbf{U}}^\top}({{\mathbf{\tilde w}}^{k}}-{{\mathbf{\tilde w}}^{*}}) \right\|$,

    \item ${{\left\| {{\mathbf{\tilde w}}^{k+1}}-{{\mathbf{\tilde w}}^{*}} \right\|}_{M}}\le \sqrt{\rho (\mathbf{T})}{{\left\| {{\mathbf{\tilde w}}^{k}}-{{\mathbf{\tilde w}}^{*}} \right\|}_{M}}$.
\end{enumerate}
\begin{proof}
Note that $g\circ h$ is non-expansive, according to Proposition~\ref{Proposition_1}, we have:
\begin{eqnarray} \label{eq:19}
{{\mathbf{\tilde w}}^{k+1}}-{{\mathbf{\tilde w}}^{*}}=C{{\mathbf{M}}^{-1}}\mathbf{U}\left( {{\mathbf{e}}^{k+1}}-{{\mathbf{e}}^{*}} \right),
\end{eqnarray}
\begin{eqnarray} \label{eq:20}
{{\left\| {{\mathbf{e}}^{k+1}} \rm{-} {{\mathbf{e}}^{*}} \right\|}^{2}} \rm{=}
{{\left\| g \rm{\circ} h({{\mathbf{e}}^{k}})\rm{-}g \rm{\circ} h({{\mathbf{e}}^{*}}) \right\|}^{2}} \rm{\le}
{{\left\| {{\mathbf{U}}^\top}({{\mathbf{\tilde w}}^{k}}\rm{-}{{\mathbf{\tilde w}}^{*}}) \right\|}^{2}}
\end{eqnarray}
Under the definition of $\mathbf{T}$, there is: ${{\left\| {{\mathbf{U}}^\top}({{\mathbf{\tilde w}}^{k}}-{{\mathbf{\tilde w}}^{*}}) \right\|}^{2}}={{\left\| \mathbf{T}({{\mathbf{e}}^{k}}-{{\mathbf{e}}^{*}}) \right\|}^{2}}$, and thus
\begin{eqnarray} \label{eq:21}
\nonumber {{\left\| {{\mathbf{e}}^{k+1}}-{{\mathbf{e}}^{*}} \right\|}^{2}}\le {{\left\| \mathbf{T}({{\mathbf{e}}^{k}}-{{\mathbf{e}}^{*}}) \right\|}^{2}},
\end{eqnarray}
Consequently, we have:
\begin{eqnarray} \label{eq:22}
\nonumber {{\left\| {{\mathbf{e}}^{k+1}} \rm{-} {{\mathbf{e}}^{*}} \right\|}^{2}} \rm{\le} {{\left( {{\mathbf{e}}^{k}}
\rm{-}{{\mathbf{e}}^{*}} \right)}^\top}\left( {{\mathbf{T}}^{2}} \right)\left( {{\mathbf{e}}^{k}}\rm{-}{{\mathbf{e}}^{*}} \right)
\rm{\le} \rho ({{\mathbf{T}}^{2}}){{\left\| {{\mathbf{e}}^{k}} \rm{-} {{\mathbf{e}}^{*}} \right\|}^{2}}.
\end{eqnarray}
By reformulating above, condition 1 can be satisfied:
\begin{eqnarray} \label{eq:23}
\left\| {{\mathbf{e}}^{k+1}}-{{\mathbf{e}}^{*}} \right\|\le \sqrt{\rho ({{\mathbf{T}}^{2}})}\left\| {{\mathbf{e}}^{k}}-{{\mathbf{e}}^{*}} \right\| .
\end{eqnarray}
Multiplying ${{\mathbf{\tilde X}}^\top}$ on both sides of (\ref{eq:19}), and combining with
(\ref{eq:20}), we obtain:
\begin{eqnarray} \label{eq:24}
\begin{aligned}
{{\left\| {{\mathbf{U}}^\top}\left( {{\mathbf{\tilde w}}^{k+1}} \rm{-}
{{\mathbf{\tilde w}}^{*}} \right) \right\|}^{2}} & \rm{=}
{{\left\| \mathbf{T}({{\mathbf{e}}^{k+1}}  \rm{-} {{\mathbf{e}}^{*}}) \right\|}^{2}} \rm{\le}
\rho ({{\mathbf{T}}^{2}}){{\left\| {{\mathbf{e}}^{k+1}} \rm{-} {{\mathbf{e}}^{*}} \right\|}^{2}}
\nonumber \\
 & \le \rho ({{\mathbf{T}}^{2}}){{\left\| {{\mathbf{U}}^\top}\left( {{\mathbf{\tilde w}}^{k}} \rm{-}
 {{\mathbf{\tilde w}}^{*}} \right) \right\|}^{2}}, \nonumber
\end{aligned}
\end{eqnarray}
which can be reformulated as:
\begin{eqnarray} \label{eq:25}
\nonumber \left\| {{\mathbf{U}}^\top}({{\mathbf{\tilde w}}^{k+1}} - {{\mathbf{\tilde w}}^{*}}) \right\| \le
 \sqrt{\rho ({{\mathbf{T}}^{2}})} \left\| {{\mathbf{U}}^\top}({{\mathbf{\tilde w}}^{k}} -
 {{\mathbf{\tilde w}}^{*}}) \right\| ,
\end{eqnarray}
and satisfies condition 2.
From (\ref{eq:19}), we have:
%
\begin{align} \label{eq:26}
\left\| {{\mathbf{\tilde w}}^{k+1}} \!\rm{-} {{\mathbf{\tilde w}}^{*}} \right\|_{M}^{2} \!\rm{=}
{{\left( {{\mathbf{e}}^{k+1}} \!\rm{-}\! {{\mathbf{e}}^{*}} \right)}^\top}\mathbf{T}\left( {{\mathbf{e}}^{k+1}}
\!\rm{-} {{\mathbf{e}}^{*}} \right) \rm{\le} \rho (\mathbf{T}){{\left\| {{\mathbf{e}}^{k+1}} \!\rm{-}
{{\mathbf{e}}^{*}} \right\|}^{2}} . \nonumber
\end{align}
Combining (\ref{eq:20}) and the definition of $\mathbf{M}$, we have:
\begin{eqnarray} \label{eq:27}
{{\left\| {{\mathbf{\tilde w}}^{k+1}} \rm{-} {{\mathbf{\tilde w}}^{*}} \right\|}_{M}} \rm{\le}\!
 \sqrt{\rho (\mathbf{T})}\left\| {{\mathbf{U}}^\top}\!({{\mathbf{\tilde w}}^{k}} \rm{-} {{\mathbf{\tilde w}}^{*}}) \right\|\rm{\le}\! \sqrt{\rho (\mathbf{T})}{{\left\| {{\mathbf{\tilde w}}^{k}} \rm{-} {{\mathbf{\tilde w}}^{*}} \right\|}_{M}} .
 \nonumber
\end{eqnarray}
Thus, the third condition 3 holds and
${{\mathbf{\tilde w}}^{k}}$ converges to ${{\mathbf{\tilde w}}^{*}}$ $\it q$-linearly \cite{allain2006global}.
\end{proof}
\end{Theorem}


\bibliographystyle{IEEEtran}
\bibliography{scf_ref}

\end{document}